\documentclass[10pt,journal,compsoc]{IEEEtran}
\ifCLASSOPTIONcompsoc
  \usepackage[nocompress]{cite}
\else
  \usepackage{cite}
\fi
\ifCLASSINFOpdf
\else
\fi

\usepackage{booktabs}
\usepackage{multirow}
\usepackage{tabularx}
\usepackage{amsmath,amsfonts}
\usepackage{algorithmic}
\usepackage{algorithm}

\usepackage{bbding}
 
\usepackage{xcolor}

\usepackage{array}
\usepackage[caption=false,font=normalsize,labelfont=sf,textfont=sf]{subfig}
\usepackage{textcomp}
\usepackage{stfloats}
\usepackage{url}
\usepackage{verbatim}
\usepackage{graphicx}
\usepackage{makecell}
\usepackage{cite}
\begin{document}

\title{FinLLMs: A Framework for Financial \\ Reasoning Dataset Generation with \\ Large Language Models}

\author{Ziqiang Yuan$^{a}$, Kaiyuan Wang$^b$, Shoutai Zhu$^{a}$, Ye Yuan$^{a}$, Jingya Zhou$^c$, Yanlin Zhu$^d$, Wenqi Wei$^{e*}$

\IEEEcompsocitemizethanks{
% %\IEEEcompsocthanksitem Work done while the authors were at Georgia Institute of Technology.
\IEEEcompsocthanksitem \indent $^{a}$Ziqiang Yuan, Shoutai Zhu and Ye Yuan are with the School of Computer Science and Technology, Beijing Institute of Technology,  Beijing, China,100081
\IEEEcompsocthanksitem \indent $^{b}$KaiYuan Wang is with Central University of Finance and Economics, Beijing, China,100081
\IEEEcompsocthanksitem \indent $^{c}$Jingya Zhou is with School of Computer Science and Technology, Soochow University, Suzhou, Jiangsu, China, 215006.
\IEEEcompsocthanksitem \indent $^{d}$Yanlin Zhu is with Moody's Analytics, New York City, NY 10007. 
\IEEEcompsocthanksitem \indent $^{e}$Wenqi Wei is with Computer and Information Science Department, Fordham University, New York City, NY 10023. 
\IEEEcompsocthanksitem \indent $^{*}$Corresponding author.
\protect \\ 
% % \indent Yanzhao Wu is with School of Computing and Information Sciences, Florida International University, Miami, FL, 33199. \protect\\ 
\noindent E-mail:   %\{wenqiwei,yanzhaowu,khchow\}@gatech.edu,
\{ziqiangy,shoutaizhu,yuan-ye\}@bit.edu.cn, caranwang@email.c-\\ufe.edu.cn, jy\_zhou@suda.edu.cn, yanlinzhu2018@gmail.com, wenqiwei@fordham.edu}
% %,  , % emregursoy@ku.edu.tr, truexs@denison.edu, khchow@gatech.edu, yawu@fiu.edu.}% <-this % stops an unwanted space 
% %\hfil\break # for a new line
\thanks{Manuscript received xxxx xx, xxxx; revised xxxxx xx, xxxx.}
}


\markboth{Journal of \LaTeX\ Class Files,~Vol.~14, No.~8, August~2015}%
{Shell \MakeLowercase{\textit{et al.}}: Bare Demo of IEEEtran.cls for Computer Society Journals}


\IEEEtitleabstractindextext{%
\begin{abstract}
Large Language models (LLMs) usually rely on extensive training datasets. In the financial domain, creating numerical reasoning datasets that include a mix of tables and long text often involves substantial manual annotation expenses. To address the limited data resources and reduce the annotation cost, we introduce FinLLMs, a method for generating financial question-answering data based on common financial formulas using Large Language Models. First, we compile a list of common financial formulas and construct a graph based on the variables these formulas employ.  We then augment the formula set by combining those that share identical variables as new elements. Specifically, we explore formulas obtained by manual annotation and merge those formulas with shared variables by traversing the constructed graph. Finally, utilizing GPT-3.5, we generate financial question-answering data that encompasses both tabular information and long textual content, building on the collected formula set.  Our experiments demonstrate that synthetic data generated by FinLLMs effectively enhances the performance of several large-scale numerical reasoning models in the financial domain, outperforming two established benchmark financial question-answering datasets.
\end{abstract}

\begin{IEEEkeywords}
Large Language Models, Question Answering, Data Generation
\end{IEEEkeywords}}

\maketitle

\section{Introduction}
\IEEEPARstart{F}{inancial} analysis is a critical means of assessing business performance. To facilitate high-quality, timely decision-making, professional analysts engage in intricate numerical reasoning, often between financial reports~\cite{chen-etal-2021-finqa}. In the pursuit of automating this process, several retriever-generator question-answering (QA) frameworks have been proposed, such as FinQA and DyRRen~\cite{Li_Zhu_Liu_Ju_Qu_Cheng_2023}. 
\begin{figure*}[t]
\centering
\includegraphics[width=0.9\textwidth]{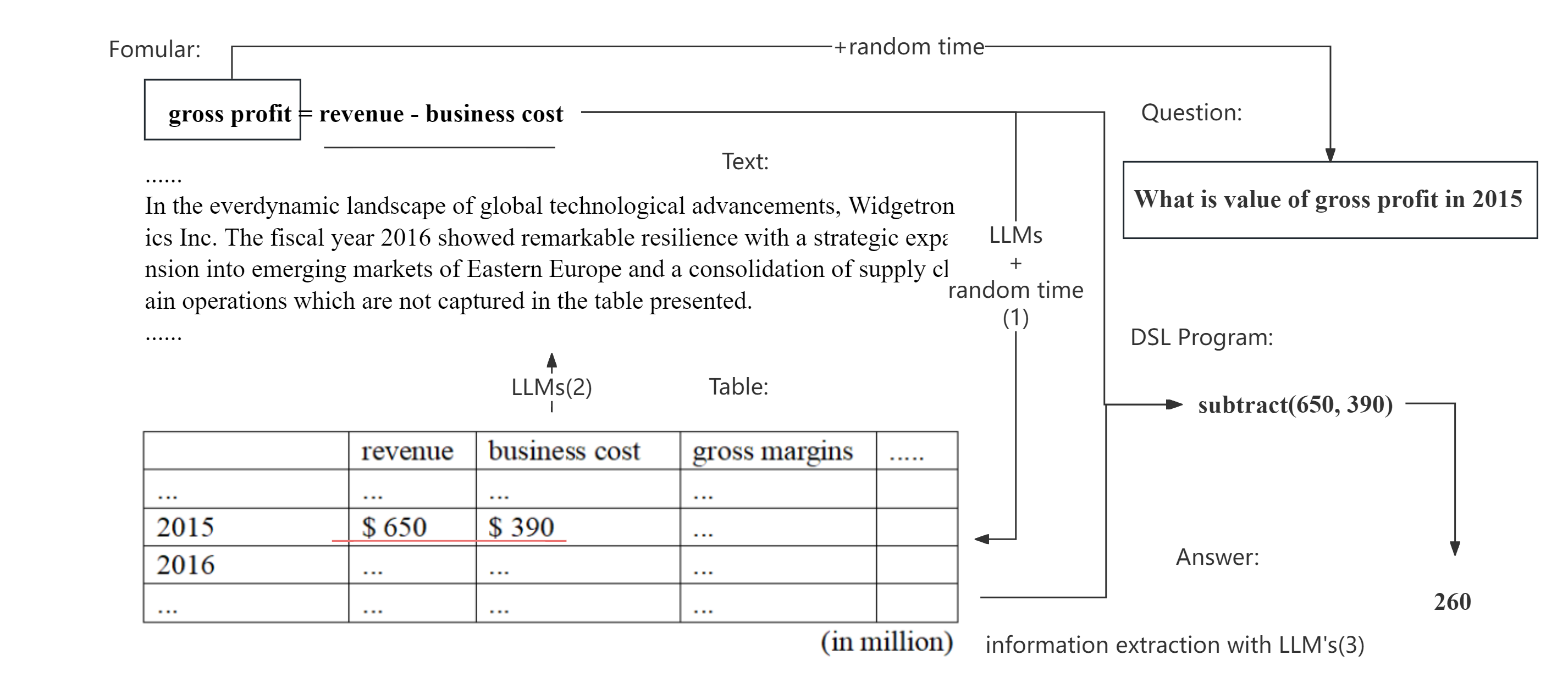} % Reduce the figure size so that it is slightly narrower than the column. Don't use precise values for figure width. This setup will avoid overfull boxes.
\vspace{-0.4cm}
\caption{Example of generating FinQA data from formulas. Throughout the entire process, the large language model is involved in three main tasks. Firstly, it combines the dependent variables of the formula with the specified time range to generate a table. Then, it extracts the values of variables from the table based on the time points relevant to the questions. Finally, it generates text related to the table. We utilize the independent variables of the formula and random time points to generate questions through templates. The ultimate answers are generated using the formula and the values extracted from the table. It is important to note that in this example, the supporting facts for the questions are sourced from the table. However, when generating supporting facts from text, we first use the large language model and a randomly chosen time range to generate text. Subsequently, we generate tables unrelated to the formula variables to avoid contradictory information.}
\label{fig10}%%%%%% need
\end{figure*}
However, constructing a question-answering model hinges on curated sets of manually labeled question-answer pairs for training. In the financial domain, this manual annotation process comes with notably higher costs~\cite{MultiHiertt}. Several factors contribute to this heightened expense: (1) Manual annotation of financial question-answering datasets demands the expertise of highly skilled labelers. This requirement stems from the need for financial domain experts, as annotators must meticulously read and perform complex numerical reasoning across multiple financial reports. (2) The content within financial question-answer pairs is intricately layered. A financial question-answering model relies on comprehending textual context and engaging in numerical reasoning specific to accounting practices. Consequently, to facilitate the training of financial numerical reasoning, experts are tasked not only with annotating the questions and answers but also with annotating the intricate financial arithmetic expressions.

The automatic generation of question-answer pairs offers a viable solution to solve the shortage of training corpus for question-answer systems. Recently, many studies in question answering have attempted to use automatically generated data to expand the training data set and improve model performance. For example, the LIQUID framework was introduced by~\cite{DBLP:conf/aaai/LeeKK23} to automatically generate list-based QA data.  However, applying these methods to the financial domain poses distinct challenges because of the unique intricacies of the question-answer annotations. The financial domain usually demands an elevated level of accuracy in numbers and arithmetic expressions.

In this paper, we present a novel framework called FinLLMs, designed for automated creation of financial question-answering datasets. Our methodology comprises three crucial steps for generating financial reasoning data. First, we collect a compilation of commonly used financial formulas, sourced from ``Principles of Accounting"~\cite {needles2013principles}. To capture the intricate relationships between formulas, where the dependent variable of one formula can exist as an independent variable in other formulas, we establish a formula graph, with formulas as nodes and their associations as edges.  We further augment the set of formulas by merging those containing the same variables as new elements. Specifically, we explore and merge those formulas by traversing the constructed graph. This strategy allows us to derive a substantial number of realistic formulas from a relatively small manually collected set. Next, on top of the formula set, we use GPT-3.5 to generate financial question-answering data containing tables and long texts. In the generated dataset, each formula corresponds to a specific type of question and calculation method. We generate multiple examples with table or text as supporting facts on the basis of each formula. 

FinLLMs offers a distinct advantage by enabling the creation of large-scale financial question-answering datasets without relying on labeled data to train the model. The process, detailed in Section~\ref{sec:3}, is illustrated in \textbf{Figure~\ref{fig10}}. Initially, we randomly generate a time range contained in the table information and the time points relevant to the questions. Subsequently, we utilize the dependent variables of the formulas, combined with the time range, to generate table information using the large language model. Simultaneously, we generate questions in the QA pair using the independent variables of the formulas. This ensures that the generated table data contains the necessary information to answer the questions. Additionally, we transform the format of the formula into a Domain-Specific Language (DSL) program comprising variables. We retrieve the values corresponding to the variables from the table and replace the variables with numerical values. Finally, we employ the large language model to generate text related to the table and calculate answers to the questions using the DSL program. By generating data based on formulas as the framework, we ensure that the generated data exhibits the following characteristics:
\begin{itemize}
    \item \textbf{Accuracy of Questions:} Since both supporting facts and questions in the generated data are derived from formulas, we can guarantee that the generated questions are answerable using the information provided in the supporting facts. This approach prevents the occurrence of invalid or unanswerable questions in the dataset.
    \item \textbf{Accuracy of Answers:} The DSL programs used in the answers are directly derived from the formulas, ensuring the correctness of the calculation process. This guarantees the accuracy of the answers provided in the generated data.
\end{itemize}
In Section~\ref{sec:4}, we evaluate and compare the proposed FinLLMs with two baseline datasets: FinQA and TAT-QA, both of which have been manually annotated. Our evaluation involves testing these datasets using four model configurations: FinQANet on BERT, FinQANet on RoBERTa-large, DyRRen on BERT, and DyRRen on RoBERTa-large. Our experiments show that FinLLMs-synthesized data consistently enhances the execution accuracy and program accuracy by at least 2\% across multiple model variants when compared with the accuracy obtained using only human-labeled data.
In Section~\ref{sec:5}, we conduct ablation studies to investigate the influencing factors on the effectiveness of the proposed methods. At last, we validate the quality of the generated data and discuss their limitations. 

In summary, our primary work encompasses three key aspects. Firstly, we introduce a novel framework for generating financial question-answering (QA) data based on financial formulas. Secondly, we leverage a graph traversal technique to augment the pool of formulas, thereby effectively reducing the need for manual annotation. Lastly, we conduct comparative and ablation experiments, demonstrating that the hybrid dataset we generated substantially enhances model performance in this domain. Through these experiments, we validate the rationality and necessity of each step in our approach.

\begin{figure*}[t]
\centering
\includegraphics[width=0.95\textwidth]{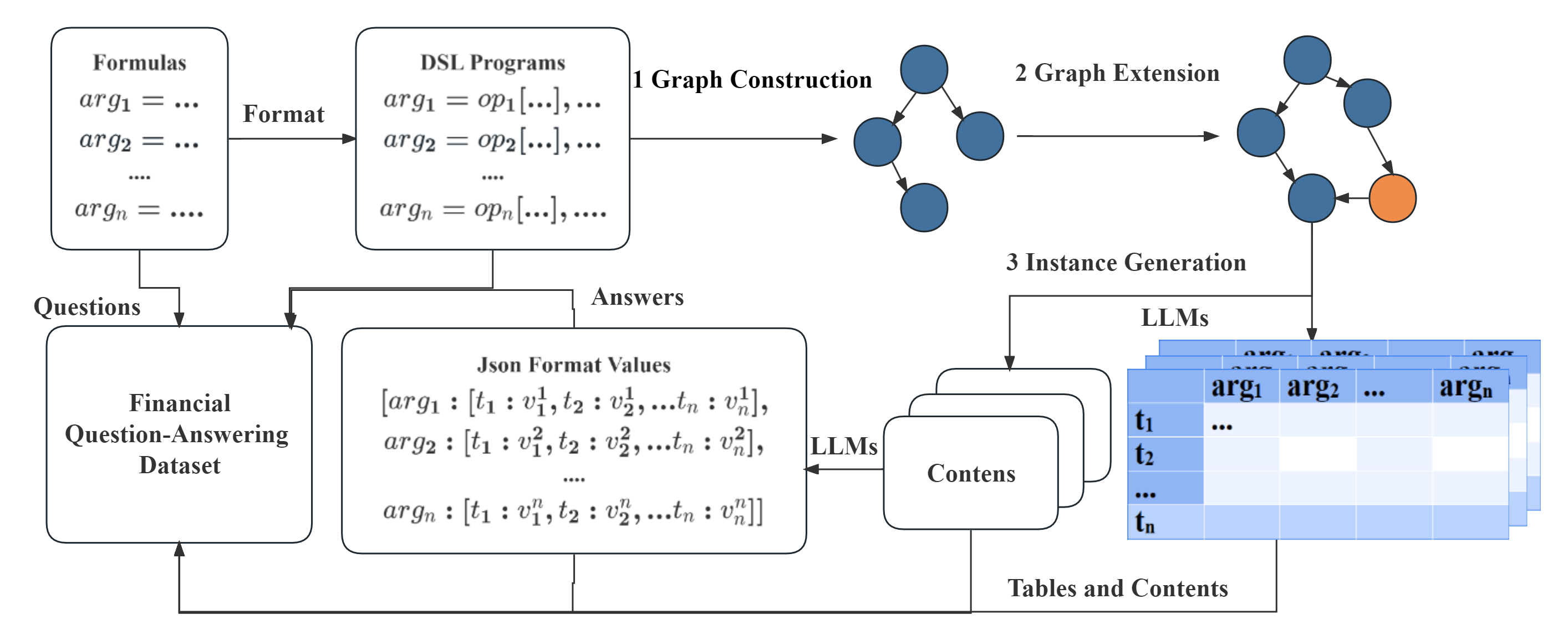} 
\vspace{-0.4cm}
\caption{Overview of FinLLMs with three steps. (1) Graph Construction: We collect standard financial formulas, format them into DSL programs and construct a graph based on the variables involved. (2) Graph Extension: We augment the set of formulas by combining those containing the same variables as new elements. Specifically, we explore and merge those formulas by traversing the constructed graph. (3) Example Generation: We use GPT-3.5 to generate financial question-answering data containing tables and long texts according to the formula set.}
\label{fig1}
\end{figure*}

\section{Problem Formulation}

\label{sec:2}

Financial reasoning is a challenging problem since it requires the system to extract relevant information across heterogeneous sources in financial reports and build a numerical reasoning path connecting all the information. These financial reports, such as the 10-K\footnote{\url{https://www.sec.gov/files/form10-k.pdf}}, 10-Q\footnote{\url{https://www.sec.gov/files/form10-q.pdf}}, and S\&P 500 earning reports, often feature a blend format of tabular data $T$ and unstructured textual data $C$. Given a problem $Q$, the task of financial question-answering is to generate a reasoning program $Prog$ utilizing both tables and unstructured text within the financial report:
\begin{equation}
Prog=\{w_1,w_2,...w_n\},
\end{equation}
where $w_i$ is the program tokens defined by Domain-Specific Language (DSL). At each timestep, the financial reasoning model makes a prediction for a $w_i$. This prediction is conditioned on previous tokens $(w_0,w_1,...,w_{i-1})$ and the input $x$, with the goal of decoding the entire operation program  $Prog=\{w_1,w_2,...w_n\}$ of length $n$:
\begin{equation}
P(Prog|T,C,Q)=\prod_{i=1}^n P(w_i|w_{<i},T,C,Q),
\end{equation}
 
During training time, the financial reasoning model minimizes the negative log-likelihood
(NLL) using the following objective:
\begin{equation}
L(\theta^{enc}, \theta^{dec}) = -logP(Prog|T,C,Q;\theta^{enc}, \theta^{dec})
\end{equation}
At test time, the financial reasoning model only observes the input text when predicting operation programs:
\begin{equation}
\hat{Prog} = argmax_{Prog}P(Prog|T,C,Q)
\end{equation}
Then, the generated reasoning program $Prog$ is executed to get the answer $A$:
\begin{equation}
P(A|T,C,Q)=\sum P(Prog_i|T,C,Q),
\end{equation}
where $\{Prog_i\}$ is the set of correct programs. For the unstructured text, we use the formula template $F$ to extract the variables from the financial report for the reasoning program $Prog$. While financial tables frequently feature a timing description header in the first row and an item description header in the first column, some tables may have more complicated layouts, e.g., nested structures. 
We focus on regular table layout and 
convert the table into sentences via table-to-text template~\cite{chen-etal-2021-finqa} during dataset generation.

%\subsubsection{.} 

In this study, we
consider DSL consisting of six mathematical operations as executable programs, namely,  
\textit{add, subtract, multiply, divide, greater, exp}. These programs are structured as a sequence of operations:
\begin{equation} 
arg_r = op_1[args_1], op_2[args_2]..., op_n[args_n].
\end{equation}
These mathematical operations accept arguments that can be either numbers extracted from the provided financial report or numerical outcomes from a preceding step.  As we convert the table into sentences, table operations, such as finding the maximum, minimum, sum, or average value of a specific row or column, can be executed through combinations of these mathematical operations. For example, calculating the average revenue over two years in a table can be reformulated as the program  $\frac{add[year1, year2]}{\# years}$, where the denominator is the number of columns involved.

\begin{figure}[t]
\centering
\includegraphics[width=1.0\columnwidth]{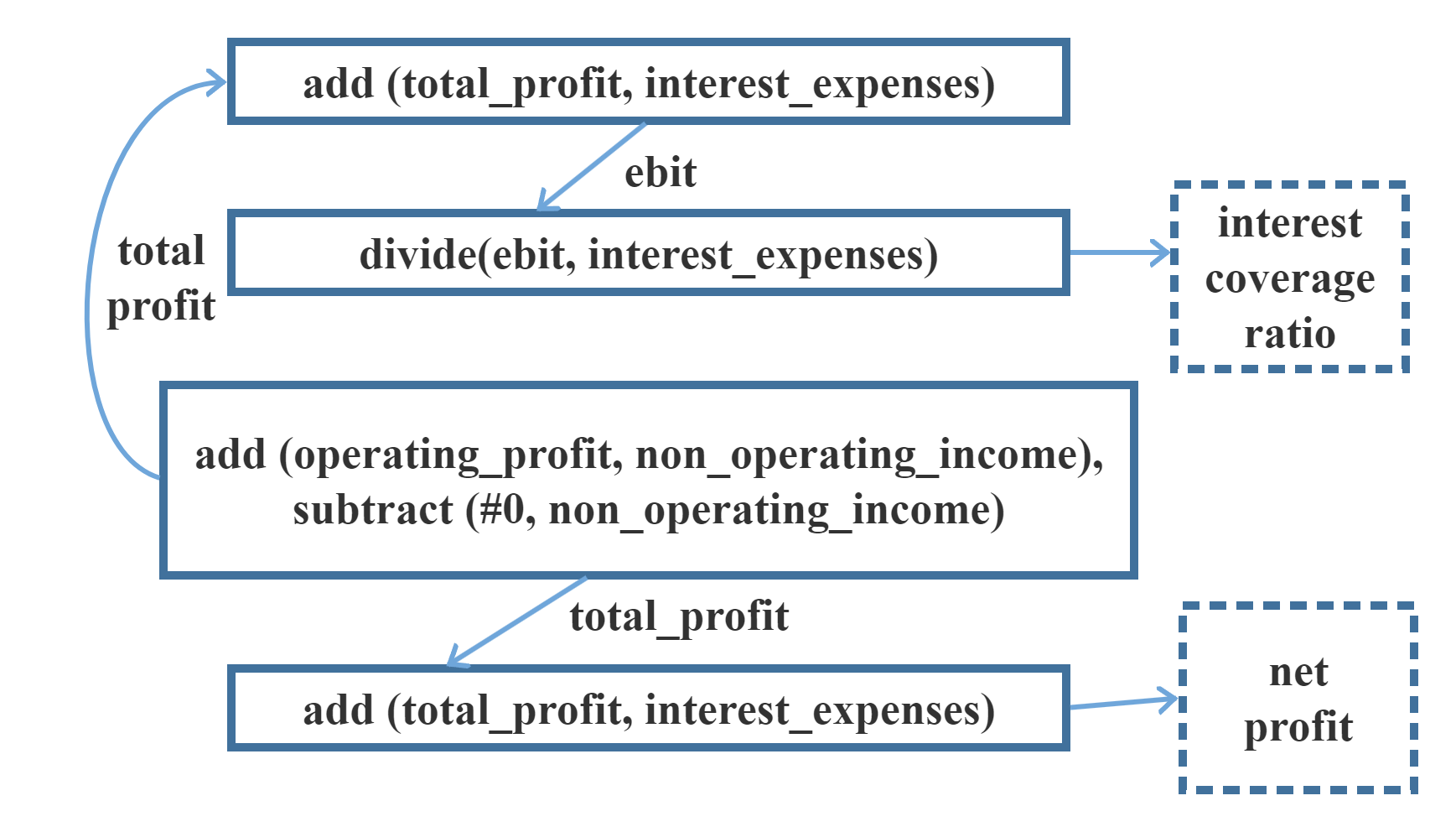} 
\vspace{-0.4cm}
\caption{This graph is composed of four formulas: ebit = total profit + interest expense, interest coverage ratio = ebit / interest expense, net profit = total profit - income tax expense, total profit = operating profit + non-operating income - non-operating expense. The nodes in the dotted box in the figure represent all the nodes whose independent variables contain the variables in the box.}
\label{fig2}
\end{figure}

\section{Financial Reasoning Dataset Generation}

\label{sec:3}

In this section, we detail the approach to financial reasoning dataset generation. 
Our goal is to automatically generate datasets $\{D_i\}$ that can be used for numerical reasoning model training in the financial domain. 
Our framework consists of three components: (1) \textbf{Graph construction} collects standard financial formulas $\{F_i\}$ from ``Principles of Accounting"~\cite {needles2013principles},
 formats them into DSL programs and constructs a graph based on the variables involved; 
(2) \textbf{Graph extension} augments the set of formulas by combining those containing the same variables; (3) \textbf{Example generation} leverages the capabilities of large language models, e.g., GPT-3.5 in our prototype, to generate financial question-answering data.  \textbf{Figures~\ref{fig1}} illustrates the financial reasoning dataset generation process.

\subsection{Graph Construction}

\label{sec:3.1}

\subsubsection{Formulating Financial Reasoning Graph.} 
We initiate the construction of the financial reasoning graph $G(V,E)$ by utilizing the variables obtained from collected standard financial formulas.  Specifically, 
we model the table and text data for numerical reasoning with formulas. Each formula corresponds to a question and DSL program. Variables within these formulas are divided into dependent variables and independent variables. Remarkably, independent variables in one formula may function as dependent variables in others. 

To establish the graph, we treat variables as nodes and the associations between variables as edges. 
Consequently, the financial reasoning graph $G(V,E)$ is composed of nodes $V={v_1,v_2,...v_n}$ and edges $E={e_1,e_2,...e_n}$. Each vertex $v_i$ represents the reasoning program $Prog_i$ derived from the formula $F_i$. Correspondingly, an edge $e_i$ is the dependent variable of the formula $F_i$, pointing to the node that takes it as the independent variable. \textbf{Figures~\ref{fig2}} presents an example consisting of four formulas.

\subsubsection{Enriching Graph Information. }

To diversify the types and increase the quantities of questions and DSL programs, we construct formula graphs by growing the financial reasoning graph. The graph information enrichment process involves node construction, edge construction, and the introduction of the time dimension.

\begin{algorithm}[t]
\caption{Graph Construction}\label{alg:alg1}
\begin{algorithmic}
\STATE
\STATE {\textsc{CONSTRUCTION}}$(\mathbf{F}):$
\STATE \hspace{0.5cm} $G=\emptyset$
\STATE \hspace{0.5cm} $\textbf{For} \, f \, \textbf{in} \, F:$
\STATE \hspace{1cm}$node = NewNode(f)$
\STATE \hspace{1cm}$G.addNode(node)$
\STATE \hspace{1cm}$G=ADDEDGE(G)$
\STATE \hspace{0.5cm} $\textbf{return} \, G$
\STATE {\textsc{ADDEDGES}}$(\mathbf{G}):$
\STATE \hspace{0.5cm}$\textbf{For} \, node1 \, \textbf{in} \, G.nodes:$
\STATE \hspace{1cm}$\textbf{For} \, node2 \, \textbf{in} \, G.nodes:$
\STATE \hspace{1.5cm} $varivables1 = node1.variables$
\STATE \hspace{1.5cm} $varivables2 = node2.variables$
% \#The dependent variable on the left side of the equal sign
\STATE \hspace{1.5cm} $variable = varivables1[0]$
% \#The independent variable on the right side of the equal sign
\STATE \hspace{1.5cm} $\textbf{if} \,  variable\, \textbf{in} \,varivables2[1:]:$
\STATE \hspace{2.0cm} $edge = NewEdge(node1,node2)$ 
\STATE \hspace{2.0cm} $\textbf{if} \,  edge\, \textbf{in} \,G.edges:$
\STATE \hspace{2.5cm} $continue$
\STATE \hspace{2.0cm} $G.addEdge(edge)$
\STATE \hspace{0.5cm} $\textbf{return} \, G$
\end{algorithmic}
\label{alg1}
\end{algorithm}

\begin{figure}[t]
\centering
\includegraphics[width=1.0\columnwidth]{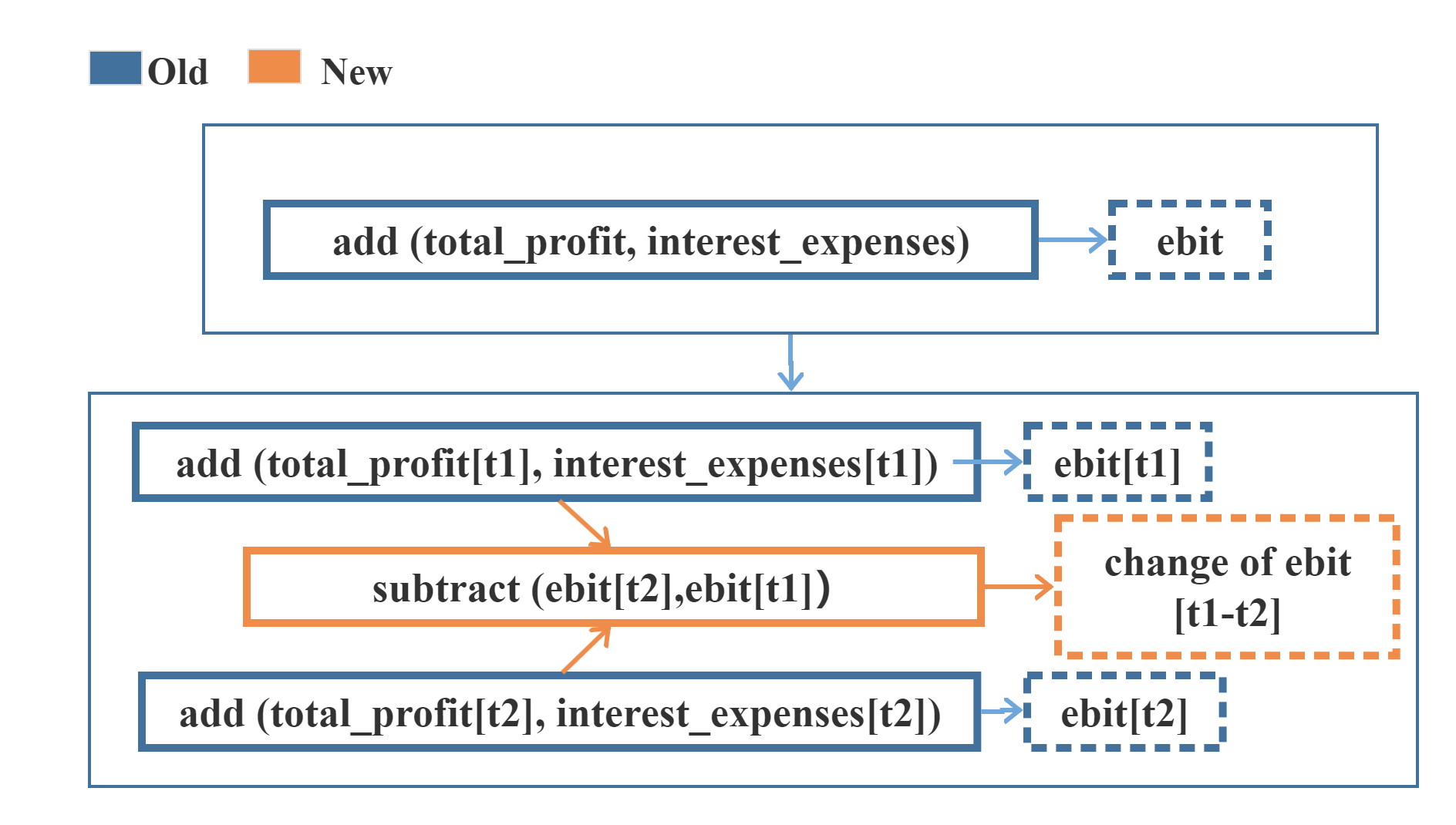} 
\vspace{-0.4cm}
\caption{We introduce the time dimension into the graph and add several formulas for each variable representing the change of variable value over time. The new formula in the figure indicates the calculation of the change of ebit between two time slices.}
\label{fig3}
\end{figure}

\textbf{ Nodes Construction.} 
We convert the collected formulas into nodes with three attributes, which are (1) $target$: the dependent variable in the formula, (2) $variables$: the set of independent variables in the formula, and (3) $program$: the DSL program of the formula. The initial formula graph we construct consists of 21 formulas and 43 variables.
 
\textbf{Edges Construction.} We then traverse all nodes and build directed edges. Specifically, when the $variables$ of node $v_i$ contain the $target$ of node $v_j$, we create an edge from node $v_j$ to node $v_i$. The detailed procedure of nodes and edges construction is outlined in \textbf{Algorithm~\ref{alg:alg1}}. The algorithm shows the traversal process of nodes and edges when they are first constructed. The method 'NewNode' represents the process of crafting a formula into a node.
 
\textbf{Time Dimension Introduction.} Considering that many financial problems need to be calculated using the values of the same variable at different times, we extended the financial reasoning graph by introducing a temporal dimension. As shown in \textbf{Figures~\ref{fig3}}, each node is transformed into two different nodes of adjacent time slices, then added to the graph, with certain specialized nodes acting as connectors. These specialized nodes are responsible for computing (1) the rate of change, (2) the change, (3) the sum, and (4) the average of each variable.

% \begin{algorithm}[H]
% \caption{Weighted Tanimoto ELM.}\label{alg:alg1}
% \begin{algorithmic}
% \STATE 
% \STATE {\textsc{TRAIN}}$(\mathbf{X} \mathbf{T})$
% \STATE \hspace{0.5cm}$ \textbf{select randomly } W \subset \mathbf{X}  $
% \STATE \hspace{0.5cm}$ N_\mathbf{t} \gets | \{ i : \mathbf{t}_i = \mathbf{t} \} | $ \textbf{ for } $ \mathbf{t}= -1,+1 $
% \STATE \hspace{0.5cm}$ B_i \gets \sqrt{ \textsc{max}(N_{-1},N_{+1}) / N_{\mathbf{t}_i} } $ \textbf{ for } $ i = 1,...,N $
% \STATE \hspace{0.5cm}$ \hat{\mathbf{H}} \gets  B \cdot (\mathbf{X}^T\textbf{W})/( \mathbb{1}\mathbf{X} + \mathbb{1}\textbf{W} - \mathbf{X}^T\textbf{W} ) $
% \STATE \hspace{0.5cm}$ \beta \gets \left ( I/C + \hat{\mathbf{H}}^T\hat{\mathbf{H}} \right )^{-1}(\hat{\mathbf{H}}^T B\cdot \mathbf{T})  $
% \STATE \hspace{0.5cm}\textbf{return}  $\textbf{W},  \beta $
% \STATE 
% \STATE {\textsc{PREDICT}}$(\mathbf{X} )$
% \STATE \hspace{0.5cm}$ \mathbf{H} \gets  (\mathbf{X}^T\textbf{W} )/( \mathbb{1}\mathbf{X}  + \mathbb{1}\textbf{W}- \mathbf{X}^T\textbf{W}  ) $
% \STATE \hspace{0.5cm}\textbf{return}  $\textsc{sign}( \mathbf{H} \beta )$
% \end{algorithmic}
% \label{alg1}
% \end{algorithm}

\begin{algorithm}[t]
\caption{Graph Extension}\label{alg:alg2}
\begin{algorithmic}
\STATE
\STATE {\textsc{EXTENSION}}$(\mathbf{G},\mathbf{steps}):$
\STATE \hspace{0.5cm} $\textbf{For} \, i \,\textbf{in}\, range(steps):$
\STATE \hspace{1cm} $newNodes=\emptyset$
\STATE \hspace{1cm} $\textbf{For} \, edge \, \textbf{in} \, G.edges:$
\STATE \hspace{1.5cm} $nodeP = edge.nodes[0]$
\STATE \hspace{1.5cm} $nodeS = edge.nodes[1]$
\STATE \hspace{1.5cm} $node = CreateNewNode(nodeP, nodeS)$
\STATE \hspace{1.5cm} $\textbf{if} \,  isValid(node)\,$
\STATE \hspace{2.0cm} $newNodes.append(node)$
\STATE \hspace{1cm}$G.addNodes(newNodes)$
\STATE \hspace{1cm}$G=ADDEDGE(G)$
\STATE \hspace{0.5cm} $\textbf{return} \, G$
\end{algorithmic}
\label{alg2}
\end{algorithm}

\subsection{Graph Extension}

We next discuss the details of graph extension. Given that the dependent variables from some formulas in certain nodes serve as independent variables in formulas within other nodes, the two nodes connected by each edge can be combined as a new candidate node as the relation of two associated nodes is expressed in the form of edges. By traversing the edges in the constructed graph, we can identify pairs of formulas that can be combined. Accordingly, we merge these pairs of formulas as new nodes and edges, and obtain new pairs of formulas to be discovered. The process of graph traversing and expanding are shown in \textbf{Algorithm~\ref{alg:alg2}}.  

\begin{figure}[t]
\centering
\includegraphics[width=1.0\columnwidth]{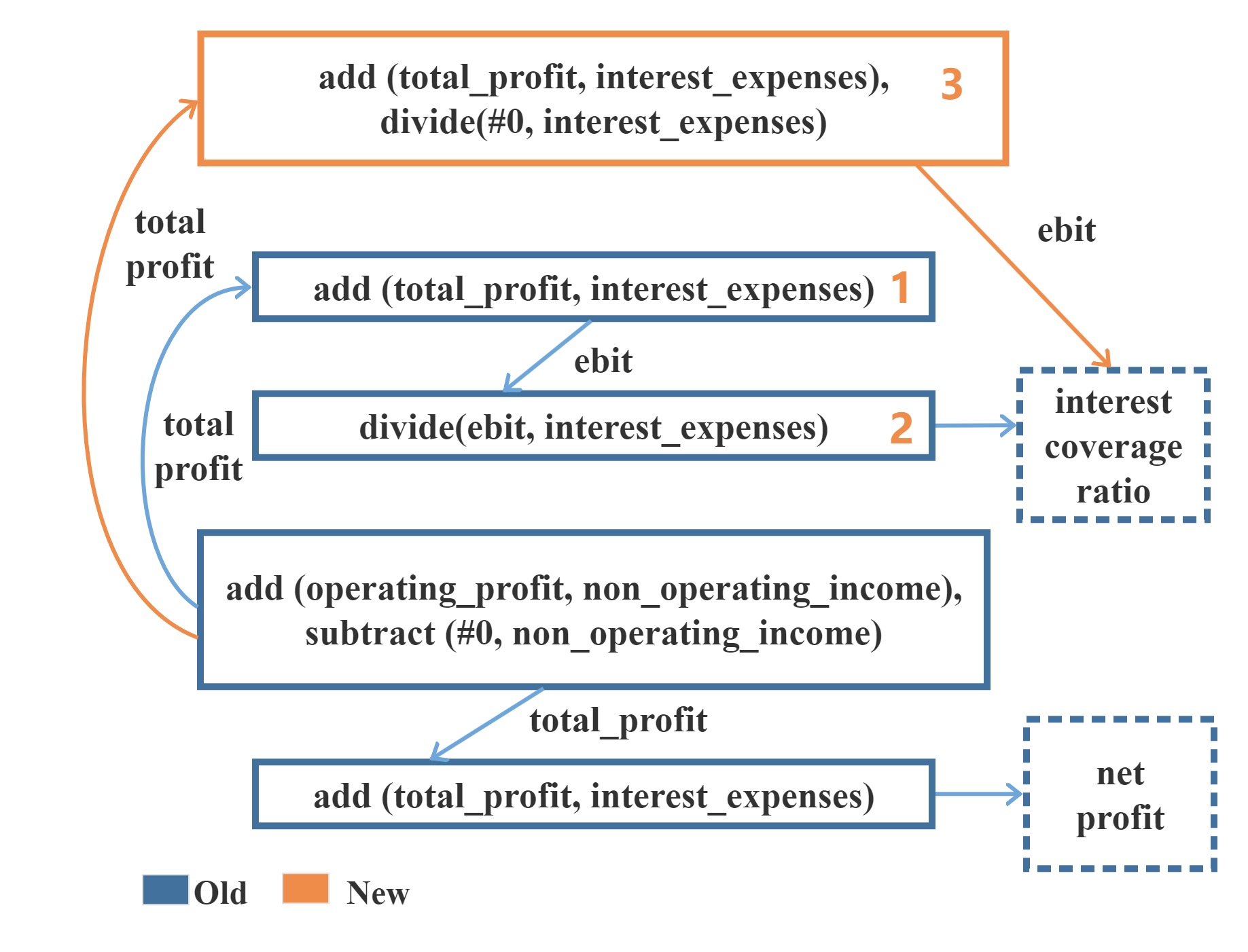} 
\vspace{-0.4cm}
\caption{The traversal process. The example in the figure does not involve time issues. When traversing the graph and going from node 1 to node 2, we can generate node 3 by combining these two nodes.}
\label{fig4}
\end{figure}

\textbf{Nodes Generation.} To generate new nodes, we first traverse edges and combine all unmarked nodes linked by directed edges. Once a pair of nodes is combined into a new one, we mark the traversed edge to prevent its reuse in subsequent traversals. The method of combining two nodes to form a new node is represented by $CreateNewNode$ in Algorithm~\ref{alg:alg2}. An illustrative example of new node generation is presented in \textbf{Figure~\ref{fig4}}.

\textbf{Nodes Filtering.} Given the complexity of the task, the newly generated nodes are filtered based on the step size limit of the DSL program and the number of independent variables. The $isValid$ method in Algorithm~\ref{alg:alg2} is used to perform this judgment. Only nodes that meet the requirements will be added to the graph.

\textbf{Nodes Addition.} We add new nodes to the graph by traversing all existing nodes and introducing directed edges in accordance with the specifications in Section~\ref{sec:3.1}.

\begin{figure}[t]
\centering
\includegraphics[width=0.85\columnwidth]{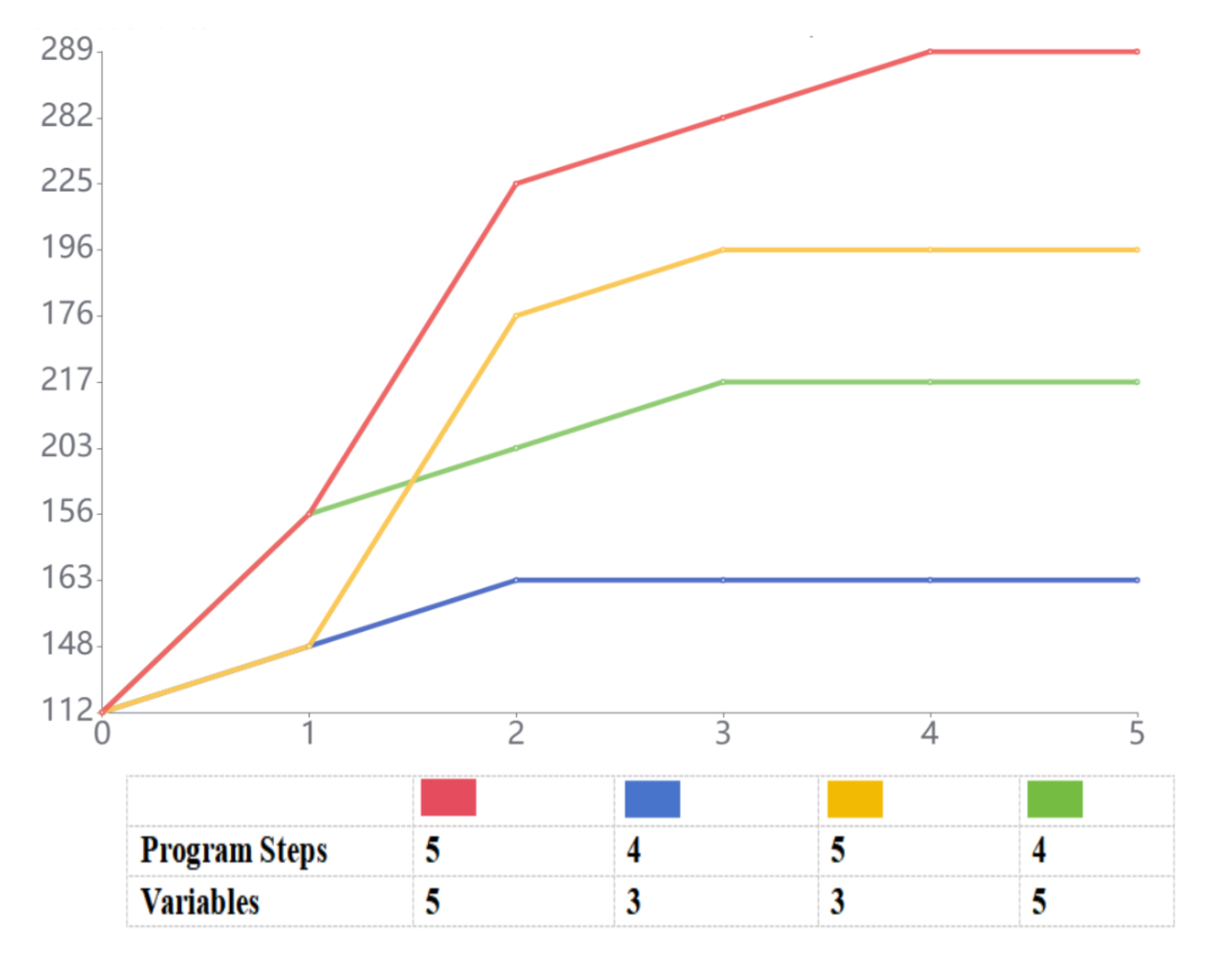} 
\vspace{-0.4cm}
\caption{The y-axis represents the number of formula nodes, and the x-axis represents the traversal steps. When the DSL program steps and the number of involved variables are limited, there exists a maximum value for the number of nodes. After reaching this value, the number of nodes in the graph will not increase with the number of traversal steps.}
\label{fig5}
\end{figure}

\textbf{Figure~\ref{fig4}} shows the process of combining two nodes to generate a new node. Empirically, we find that repeating the above traversal process five times provides us with the highest performance gain. As the traversal step increases, the newly generated nodes usually exhibit more DSL program steps and variable numbers. To balance between the processing capability of the model and the complexity of the dataset, we filter the newly generated nodes from two dimensions: (1) the number of DSL program steps and (2) the number of variables. \textbf{Figures~\ref{fig5}} shows the growth in the number of nodes under different constraints. It can be seen that the more the number of program steps and variables, the more nodes are generated in each traversal.


Note that the arguments in the reasoning program remain variable names rather than values in the graph construction and graph extension. To feed appropriate values into the corresponding variables in the reasoning program $Prog$ and question $Q$, we obtain $C$ and $T$ values from the financial report in the third step, detailed in the next section.

\subsection{Example Generation}
\begin{algorithm}[t]
\caption{Generate Financial Data}\label{alg:alg3}
\begin{algorithmic}
\STATE
\STATE {\textsc{Generate}}$(\mathbf{F}):$
\STATE \hspace{0.5cm} $T=\emptyset$
\STATE \hspace{0.5cm} $\textbf{For} \, formula \, \textbf{in} \, formulas:$
\STATE \hspace{1cm}$Time = GenerateTime(formula)$
\STATE \hspace{1cm}$Table = GenerateTable(formula,Time)$
\STATE \hspace{1cm}$TableText = GenerateTableText(Table)$
\STATE \hspace{1cm}$Text = GenerateText(formula,Time)$
\STATE \hspace{1cm}$TextTable = GenerateTextTable(Text)$
\STATE \hspace{1cm}$T.add(Table,TableText)$
\STATE \hspace{1cm}$T.add(Text,TextTable)$
\STATE \hspace{0.5cm} $\textbf{return} \, T$
\end{algorithmic}
\label{alg3}
\end{algorithm}

In the last step, we gather the formula set $\{F_i\}$ in alignment with the nodes in the expanded reasoning graph $G$, and then draw upon this set $\{F_i\}$ to obtain the table and text content from the financial report. Next, we convert these information into structured data $\{D_i\}$ that can be used for model training.

\subsubsection{Content and Table Generation.} 
We first extract the variables and time intervals that need to be included from the financial report through predefined formulas. Then, we generate financial reports using GPT-3.5 with a prompt to restrict the output format. 
\textbf{Algorithm~\ref{alg3}} provides a detailed breakdown of the steps involved. For each formula within the formula library, we generate financial reasoning data from financial reports through the following steps:
\begin{itemize}
\item  \textbf{Step 1. GenerateTime.} We use the GenerateTime function to generate a specific time range relevant to each formula.

\item  \textbf{Step 2. GenerateTable.} This step calls the GenerateTable function, which interfaces with GPT-3.5 to generate the corresponding financial table based on the formula and time.

\item  \textbf{Step 3. GenerateTableText.} We call the GenerateTableText function that harnesses the GPT-3.5 interface to generate relevant text associated with the Table generated in the previous step. Note that the key information is concentrated in the table.

\item  \textbf{Step 4. GenerateText.} The GenerateText function is utilized to invoke GPT-3.5 and create corresponding financial text content for the formula and time. 

\item  \textbf{Step 5. GenerateTextTable.} We call the GenerateTextTable function, which interfaces with the GPT-3.5, to generate relevant tables corresponding to the text generated in the previous step. Here, the key information is embedded in the text.
\end{itemize}
By executing these steps for each formula, we obtain a diverse set of financial data that encompasses various perspectives.

\subsubsection{Knowledge Extraction.} 
The generated financial reports are re-entered into GPT-3.5, which allows to extract values of variables at different time intervals. For each formula $F_i$, these obtained values are subsequently used to replace the time-related variables in question $Q_i$ and the variable names in the reasoning program $Prog_i$. 

\subsubsection{Input Formatting.}
To ensure that the generated data can accommodate the input requirement of several existing financial question-answering models~\cite{chen-etal-2021-finqa,TAT-zhu-etal-2021-tat,Li_Zhu_Liu_Ju_Qu_Cheng_2023}, we need to reformat the data through the following steps.
\begin{itemize}
\item  \textbf{Step 1. Result Calculation.} We calculate the result of each example based on the DSL program and the data obtained in the previous step.
\item \textbf{Step 2. Generation of Supporting Facts.} Sentences containing numbers that appeared in questions or formulas are set as supporting facts. 
\item \textbf{Step 3. Table Transposition.} Since we use a template to convert table rows into text, the table columns must be labelled with time, and the table rows must correspond to variables. In cases where the tables don't adhere to this format, we transpose them to make them compliant.
\end{itemize}

\begin{table*}[t]
\centering
\caption{Results of FinLLMs on different models. In the table, EA represents execution accuracy and PA represents program accuracy. TagOp is the baseline method of TAT-QA. Since the test set is not public, the table shows its accuracy in processing arithmetic-related problems on the dev dataset. Except for the 2.52$\%$ wrong examples in the dataset, the human expert performance execution accuracy of the generated part of the FinLLMs dataset was tested to be 100$\%$.}
%\resizebox{.95\columnwidth}{!}{
\begin{tabular}{l*{6}{c}}
  \toprule
  \multirow{2}*{Model} & \multicolumn{2}{c}{FinLLMs} & \multicolumn{2}{c}{FinQA} & \multicolumn{2}{c}{TAT-QA}\\
  \cmidrule(lr){2-7}
  & EA & PA & EA & PA & EA &PA \\
  \midrule
  FinQANet(BERT-base) & \textbf{53.01} & \textbf{51.09} & 50.00 & 48.00& 51.96 & 50.13\\
  FinQANet(RoBERTa-large) & \textbf{63.56} & \textbf{60.68} & 61.24 & 58.86 & 61.55 & 59.02\\
  \midrule
  DyRRen(BERT-base) & \textbf{59.55} & \textbf{57.54} & 59.37  & 57.54 & 59.02 & 57.45\\
  DyRRen(RoBERTa-large) & \textbf{65.39} & \textbf{63.38} & 63.30 & 61.29  &64.25  &61.38\\
  \midrule
TagOp(RoBERTa-large) & 47.21 & - & 57.89 & -  &45.82  &-\\
  \midrule
  Human Expert Performance & 100.00$^*$ & - & 91.16 & - & - & -\\
  \bottomrule
\end{tabular}
\label{table1}
\end{table*}

\section{Experiment}

\label{sec:4}

\subsection{Datasets}
To evaluate the performance of FinLLMs, we consider two datasets that involve generating arithmetic expressions for numerical reasoning on tabular and textual data as the baseline: FinQA~\cite{chen-etal-2021-finqa} and TAT-QA~\cite{TAT-zhu-etal-2021-tat}. FinQA comprises 8,281 financial questions collected from financial reports of S$\&$P 500 companies. These questions are split into 6,251 (75$\%$) for training, 883 (10$\%$) for development, and 1,147 (15$\%$) for testing. In TAT-QA, only a portion of the data includes arithmetic expressions for numerical reasoning. We extract, filter, and format this part of the data, resulting in 5,167 examples. We combine this with the FinQA dataset, forming a dataset for our experiments. This merged dataset is divided into 10,126 (75$\%$) for training, 1,399 (10$\%$) for development, and 1,923 (15$\%$) for testing and comparison.

The FinLLMs dataset combines the synthesized data following Figures~\ref{fig1} with the FinQA dataset and the qualified portion of the TAT-QA dataset for financial reasoning data augmentation. Therefore, the FinLLMs dataset comes in various versions, differing in size. The specific version employed for the experiments listed in \textbf{Table~\ref{table1}} comprises a total of 15,361 examples. This dataset includes 6,676 examples that require textual information to answer and 8,685 examples that require tabular information for answering. We split these examples according to the same 75/10/15 ratio as the previously mentioned datasets. The dataset and code are publicly available\footnote{\url{https://github.com/ziqiangyuan/FinLLMs}}.

\subsection{Implementation Detail}

We conduct comparative experiments on two retriever-generator models designed explicitly for numerical reasoning over tabular and textual data: FinQANet~\cite{chen-etal-2021-finqa} and DyRRen~\cite{Li_Zhu_Liu_Ju_Qu_Cheng_2023} on the three datasets: FinQA, TAT-QA, and FinLLMs. As reported in the corresponding papers, we consider BERT-base-uncased~\cite{DBLP:conf/naacl/DevlinCLT19} version and RoBERTa-large~\cite{DBLP:journals/corr/abs-1907-11692} version for FinQANet and DyRRen, respectively.  For BERT-base-uncased, the number of hidden layers is set to 12, hidden units to 768, and attention heads to 12. For RoBERTa-large, the number of hidden layers is set to 24, hidden units to 1024, and attention heads to 16.
  
When training FinQANet, we adopt the Adam optimizer with learning rate = 2e - 5 for BERT and learning rate = 1e - 5 for RoBERTa. We set batch size = 16 and max sequence length = 256. When training DyRRen, we use the Adam optimizer with learning rate = 2e - 5 for BERT and learning rate = 1.5e - 5 for RoBERTa. We set batch size = 16, max sequence length = 256, epoch = 300, and seed = 8. We set k = 3 in DyRRen's retriever.

Models of the retriever-generator class are typically divided into two components: retriever and generator. The retriever is responsible for scoring each sentence in the text and each row in the table to determine its relevance to the given question. On the other hand, the generator takes the $N$ most relevant sentences (or rows in the table) along with the question as input and produces a DSL program that can be used to compute the final answer. In the experiments, $N$ is set to 3.

The experiments are conducted on NVIDIA A100 (40GB) GPUs running
Ubuntu 20.04. Our implementations are based on Python 3.10 and we develop the deep learning models using with PyTorch 1.11. % and we select models on the development set.

%\subsection{Financial Reasoning Model Benchmarks}

\subsection{Evaluation Metrics}
Following the literature~\cite{chen-etal-2021-finqa}, we measure program accuracy (PA), i.e., proportion of correctly generated arithmetic expressions, and execution accuracy (EA), i.e., proportion of correct final answers. 
It's important to note that there are instances where an induced expression might differ from the ground truth, yet their answers are equivalent. Consequently, the execution accuracy is always at least as high as the program accuracy.

\subsection{Results}

We evaluate the performance of FinLLMs on two benchmark models: FinQANet and DyRRen, under two backbones: BERT-base-uncased and RoBERTa-large. Specificially, we first train each model using TAT-QA, FinQA, and FinLLMs datasets, respectively. Then, we test all models using the FinQA test dataset to verify and compare the contributing effect of the corresponding training dataset. Table~\ref{table1} shows the results and we make four observations. (1) On the FinQA test set, FinQANet(BERT-base), FinQANet(RoBERTa-large), and DyRRen(RoBERTa-large) models trained on FinLLMs dataset(15k version) outperform the same models trained on FinQA dataset by at least 2$\%$. (2) The DyRRen(BERT-base) model trained on the FinLLMs dataset is comparable to the model DyRRen(BERT-base) trained on the FinQA dataset as well as on the TAT-QA dataset. (3) The FinQANet(BERT-base) and FinQANet(RoBERTa-large) models trained on the FinLLMs dataset outperformed the model trained on the TAT-QA dataset by at least 1$\%$. DyRRen(RoBERTa-large) models trained on the FinLLMs dataset outperformed the model trained on the TAT-QA dataset by at least 2$\%$. 
(4) When the TapOp model is trained using FinLLMs data transformed into the TATQA format, it achieved over a 1\% improvement in performance on arithmetic-type questions compared to the original TAT-QA dataset.
These results indicate that FinLLMs-synthesized data can provide training quality similar to traditional expert-labeled datasets and even witness the solid improvement on model performance.

\begin{table}[t]
\centering
\caption{Results of Datasets on LLMs}
%\resizebox{.95\columnwidth}{!}{
\begin{tabular}[c]{ccc}
    \toprule
    & FinQA(EA) & FinLLMs(EA)  \\
    \midrule
    VICUNA-33b & 10.81  &  27.02 \\
    GPT-3.5 & 13.78 & 28.07 \\
    GPT-4 & 27.99 & 46.87 \\
    \bottomrule
\end{tabular}
\label{table2}
\end{table}

In addition, we test the performance of FinLLMs-synthesized data on state-of-the-art large language models, including GPT-3.5, GPT-4\footnote{\url{https://platform.openai.com/docs/api-reference/introduction}} and VICUNA-33b\footnote{\url{https://chat.lmsys.org/}} using one-shot learning. We compare the results with the human-label data from FinQA. \textbf{Table~\ref{table2}} shows the results. 
Our observation reveals that even the most advanced large language models are still inferior to the traditional retriever-generator model in dealing with numerical reasoning over tabular and textual data (recall Table~\ref{table1}). Furthermore, it's worth noting that the performance of synthetic data generated by FinLLMs surpasses that of human-labeled data sourced from FinQA. This outcome suggests that the synthetic data produced by FinLLMs not only exhibits a higher level of comprehension but is also more readily accessible.

These observations underscore the potential of FinLLMs to generate high-quality synthetic data, which could be a valuable asset in various applications within the financial domain. The unique approach of FinLLMs in processing and generating synthetic data yield coherent and particularly well-suited information for comprehension by the large language models, potentially revolutionizing data generation in various applications within the financial sector.

\subsection{Ablation Study}

\begin{table}[t]
\centering
\caption{ Results of FinLLMs Datasets in different traversal steps. All
tests in this table are performed on the FinQANet(BERT-
base) model. All traversal step tests in this table are performed on the FinQANet(BERT-base) model using FinLLMs 15k version.}
%\resizebox{.95\columnwidth}{!}{
\begin{tabular}[c]{ccc}
    \toprule
    \multirow{2}*{Datasets}& \multicolumn{2}{c}{FinQANet(BERT-base)}  \\
    & EA & PA  \\
    \midrule
    FinQA & 50.00 & 48.00 \\
    \midrule
    FinLLMs&&\\
    Traversal Steps 0 & 50.48 & 48.91 \\
    Traversal Steps 1 & 50.39 & 48.47\\
    Traversal Steps 2 & 51.17 & 49.52\\
    Traversal Steps 3 & 51.96 & 49.96 \\
    table-only &50.65 & 49.17\\
    text-only & 50.74 & 49.00\\
    \bottomrule
\end{tabular}

\label{table3}
\end{table}

\subsubsection{Supporting facts types} 
In FinLLMs, we categorize examples into two distinct types: 
 questions only require the information in the text to answer and questions only require the information in the table. In this experiments, we separate these two kinds of examples from FinLLMs (6,676 text-only examples and 8685 table-only examples) and integrate them with the corresponding categories in FinQA, respectively. These combined datasets are tested using the FinQANet(BERT-base) model. \textbf{Table~\ref{table3}} shows the results. We observe that the performance of table-only and text-only is comparable, and both exhibit lower performance than mixed datasets of similar scale.

\subsubsection{Number of graph traversal iterations}

Considering that the number of graph traversals iterations may impact the distribution of the generated formulas and thus affect the experimental results, we compare datasets generated with different numbers of traversal times. As shown in Figure~\ref{fig5}, when the number of DSL program steps is fewer than 4 and the number of variables is fewer than 5, the number of formulas ceases to increase after the third traversal. Therefore, we compare the four cases where the number of traversal steps is from 0 to 3. Table~\ref{table3} shows the results. Our observation is that the performance of the model gradually increases with the  traversal steps.

\subsubsection{Data size}
We next analyze the variation in the question-answering performance on FinQANet(BERT-base) with different sizes of the synthetic data.
\textbf{Table~\ref{table4}} shows that the performance tended to increase and then decrease as the data size increases, indicating the presence of optimal data size. In addition, models trained with synthetic data outperformed models trained with human-labeled data from FinQA at all data sizes. %The experimental results of the FinQA dataset and the FinLLMs 15k version Table~\ref{table4} are taken from Table~\ref{table1}. 

\begin{table}[t]
\centering
\caption{Results of FinLLMs Datasets in different sizes. All tests in this table are performed on the FinQANet(BERT-base) model.}
%\resizebox{.95\columnwidth}{!}{
\begin{tabular}[c]{ccc}
    \toprule
    \multirow{2}*{Datasets}& \multicolumn{2}{c}{FinQANet(BERT-base)}  \\
    & EA & PA  \\
    \midrule
    FinQA & 50.00 & 48.00 \\
    \midrule
    FinLLMs&&\\
    5k & 51.96 & 49.96 \\
    10k & 51.53 & 50.04\\
    15k & 53.01 & 51.09\\
    20k & 53.27 & 51.44 \\
    25k & 52.57& 50.65\\
    \bottomrule
\end{tabular}

\label{table4}
\end{table}

\subsubsection{Model and Prompt Method} During the data generation process, we have experimented with a variety of state-of-the-art large language models, e.g., GPT-3.5, GPT-4, LLaMA2-13B. In this expeirment, we consider different Prompt methods including zero-shot, one-shot and few-shot techniques. Given that the quality of data generated by LLaMA2 cannot meet the training requirements of the FinQA model, we measure the performance of FinLLMs on GPT-3.5 and GPT-4 with different prompt techniques. Specifically,  we randomly select examples from the FinQA training data as prompts and feed them to these large language models. \textbf{Table~\ref{table5}} shows the results. We observe that GPT-3.5 and GPT-4 have similar performance. However, the employment of Prompt learning, particularly the few-shot method, enabled synthetic datasets to yield enhanced model performance.
\begin{table}[t]
\centering
\caption{Performance of different LLMs and prompt learning methods when generating data. Here, the model we use is FinQANet (bert-base). In few-shot and one-shot experiments, we randomly select prompt data from the training dataset. Among them, Few-shot uses four sets of QA data as prompts.}
%\resizebox{.95\columnwidth}{!}{
\begin{tabular}[c]{ccc}
    \toprule
    \multirow{2}*{Model and Prompt Method}& \multicolumn{2}{c}{FinQANet(BERT-base)}  \\
    & EA & PA  \\
    \midrule
    GPT-3.5&&\\
     zero-shot & 53.70 & 51.35 \\
     one-shot & 53.87 & 51.70\\
     few-shot & 54.31 & 52.22\\
    \midrule
    GPT-4&&\\
     zero-shot & 53.70 & 51.35 \\
     one-shot & 53.97& 51.79\\
     few-shot & 53.62& 51.09\\
    \bottomrule
\end{tabular}

\label{table5}
\end{table}

\section{Further Analysis and Discussions}

\label{sec:5}

\subsection{Data Distribution}
\subsubsection{Supporting facts}
\begin{figure}[t]
\centering
\includegraphics[width=0.90\columnwidth]{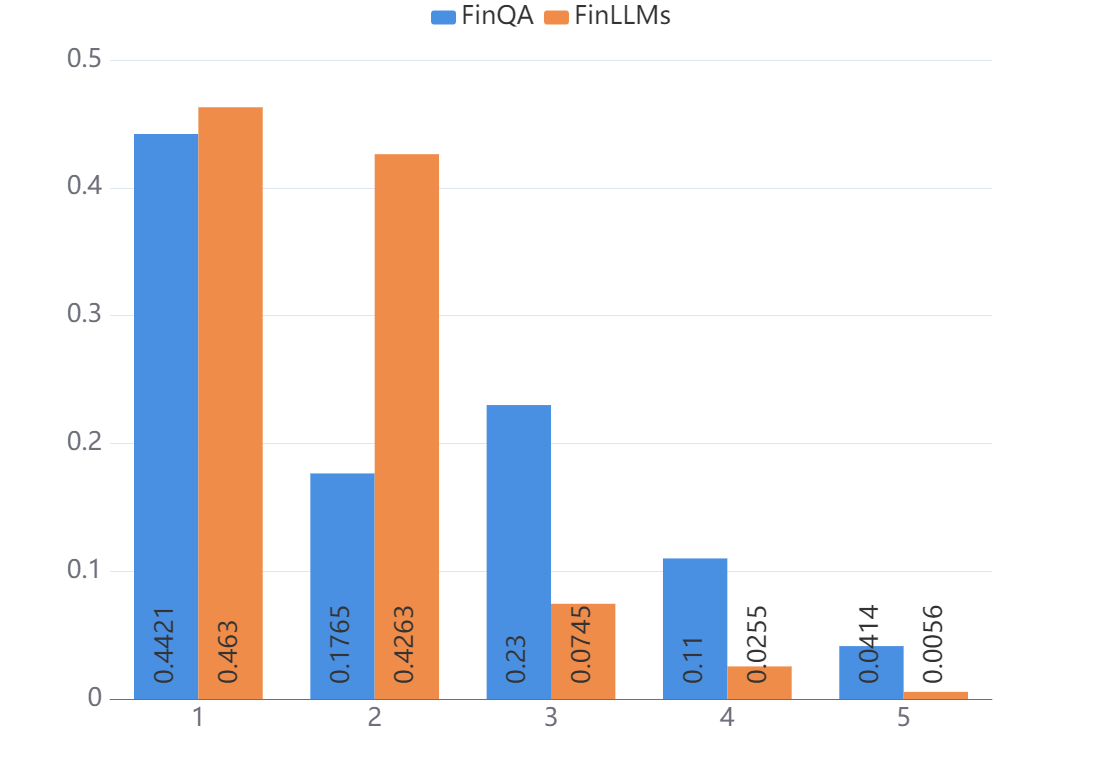} % Reduce the figure size so that it is slightly narrower than the column. Don't use precise values for figure width. This setup will avoid overfull boxes.
\vspace{-0.4cm}
\caption{Statistics of supporting facts. The blue color represents the number of statements(or rows in the table) related to each question in the original FinQA dataset. Orange represents the data distribution of the FinLLMs dataset.}
\label{fig8}
\end{figure}

As shown in \textbf{Figures~\ref{fig8}}, 44.21$\%$ of the examples in the FinLLMs-synthesized dataset rely on a single sentence or one table row as their fact, while 17.65$\%$ require two pieces of facts, 23.00$\%$ have three pieces of facts, and 15.14$\%$ have more than three pieces of facts. By comparison, the human-labeled FinQA dataset presents a different distribution, with only 46.3\% of examples having one sentence or one table row as the fact, 42.63\% requiring two pieces, 7.45\% involving three pieces, and 3.11\% having more than three pieces of facts. The complex composition of the supporting facts in the human-labeled FinQA dataset may become noise factors when generating problems. disparity in the composition of supporting facts between the FinLLMs-synthesized dataset and the human-labeled FinQA dataset could potentially explain the performance gap observed between the two (Recall Table~\ref{table1}). %%%% pick one from above to explain.

\subsubsection{DSL program steps}
Within the FinLLMs-synthesized dataset, there is a constraint that limits the number of DSL program steps to a maximum of 4. This constraint results in a distribution where approximately 45.18\% of the examples consist of a single DSL program step, 45.70\% involve two DSL program steps, 4.45\% entail three DSL program steps, and 4.67\% encompass four DSL program steps.
In general, it can be observed that examples with fewer DSL program steps tend to be simpler. This aspect contributes to our dataset having lower complexity when compared to the FinQA dataset.

\subsection{Case Study}
 The example in \textbf{Table~\ref{table6}} shows how data generated by formulas can help the model understand new knowledge and correctly answer related questions. This particular example is sourced from the FinQA dataset, identified by the ID $CDW/2017/page\_38.pdf-1$. This example is used as the test set in both training scenarios depicted in the table. In this example, the model requires knowledge of the Formula $gross\_margins=gross\_profit/revenue$ to solve the problem. Without the use of synthetic data during training, the model lacks the necessary information to determine which variables are essential to solve the problem. Consequently, the model's retriever selects the incorrect data. Furthermore, when making the final prediction, the model misplaces ``gross profit" with in the equation, incorrectly positioning it in the denominator when it should be in the numerator.
 In contrast, the model trained using the FinLLMs-synthesized dataset not only correctly applies the  knowledge of formulas to provide accurate predictions but also successfully interpreted ``net sales" as equivalent to ``revenue."

\subsection{Error Analysis}
We further extract and analyze 357 synthetic question-answering examples and discover that 97.48$\%$ of them are correct. This outcome aligns with our expectations since the programs and answers for all these examples are derived from specific formulas. However, errors in numerical calculations can arise due to large language models  overlooking numerical units, such as ``millions", during knowledge extraction from text. These errors, while minor, highlight the challenges associated with precisely interpreting and processing numerical information in text when generating synthetic examples using language models.

\begin{figure}[t]
\centering
\includegraphics[width=1.0\columnwidth]{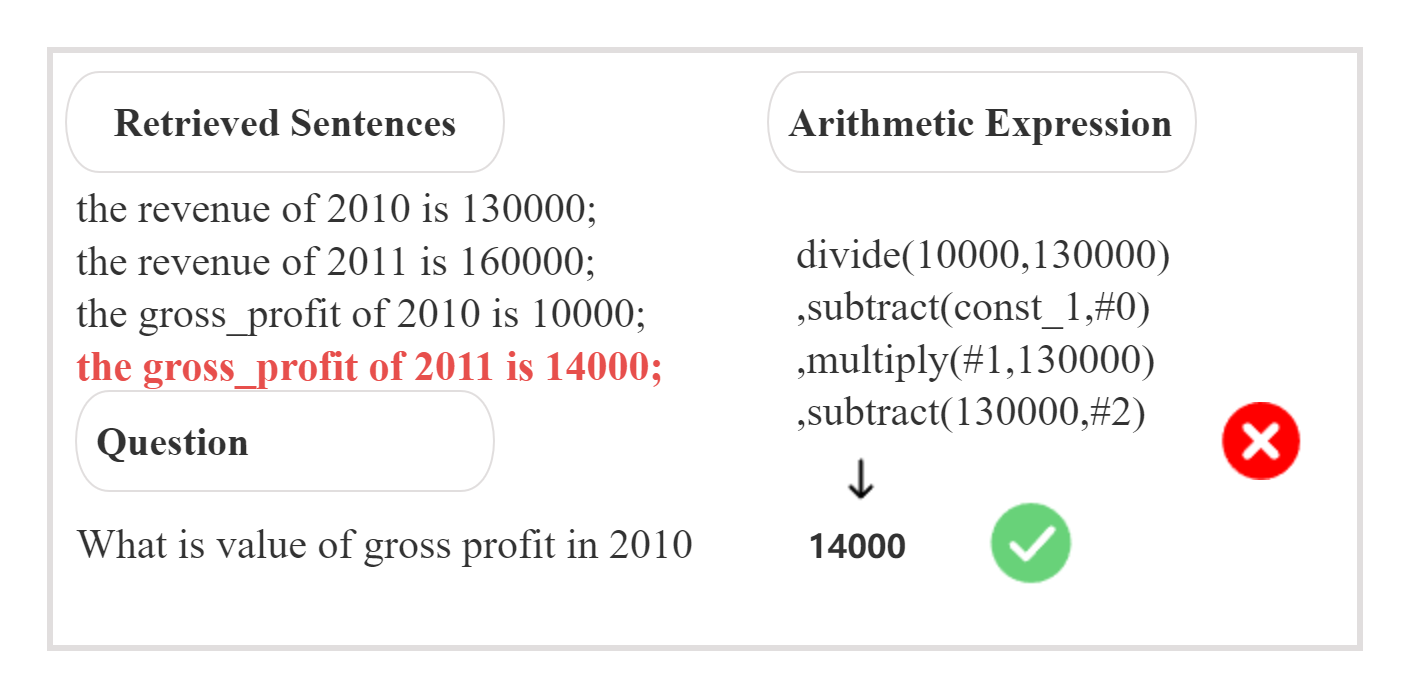} 
\vspace{-0.4cm}
\caption{Example of error: The answer to the question is in the text. When the model generates the answer, it should directly give the numerical value instead of calculating it through other variables.}
\label{fig7}
\end{figure}

Additionally, we find some anomalies in the examples that are not errors. 
As shown in \textbf{Figures~\ref{fig7}}, 
32.7$\%$ of the examples exhibit excessively complicated calculation processes. In these cases, simpler calculation methods can be employed based on known conditions, which would also allow for the simplification of the corresponding DSL programs as well. 

% \begin{table*}[t]
% \centering
% \caption{Results of FinLLMs on different models. In the table, EA represents execution accuracy and PA represents program accuracy. TagOp is the baseline method of TATQA. Since the test set is not public, the table shows its accuracy in processing arithmetic-related problems on the dev dataset. Except for the 2.52$\%$ wrong examples in the dataset, the human expert performance execution accuracy of the generated part of the FinLLMs dataset was tested to be 100$\%$.}
%\resizebox{.95\columnwidth}{!}{
% \begin{tabular}{l*{6}{c}}
%   \toprule
%   \multirow{2}*{Model} & \multicolumn{2}{c}{FinLLMs} & \multicolumn{2}{c}{FinQA} & \multicolumn{2}{c}{TAT-QA}\\
%   \cmidrule(lr){2-7}
%   & EA & PA & EA & PA & EA &PA \\
%   \midrule
%   FinQANet(BERT-base) & \textbf{53.01} & \textbf{51.09} & 50.00 & 48.00& 51.96 & 50.13\\
%   FinQANet(RoBERTa-large) & \textbf{63.56} & \textbf{60.68} & 61.24 & 58.86 & 61.55 & 59.02\\
%   \midrule
%   DyRRen(BERT-base) & \textbf{59.55} & \textbf{57.54} & 59.37  & 57.54 & 59.02 & 57.45\\
%   DyRRen(RoBERTa-large) & \textbf{65.39} & \textbf{63.38} & 63.30 & 61.29  &64.25  &61.38\\
%   \midrule
% TagOp(RoBERTa-large) & 47.21 & - & 57.89 & -  &45.82  &-\\
%   \midrule
%   Human Expert Performance & 100.00$^*$ & - & 91.16 & - & - & -\\
%   \bottomrule
% \end{tabular}

% \label{table1}
% \end{table*}

\begin{table*}[t]
\centering
\caption{Case Study. This example shows how data generated by formulas can help the model understand new knowledge and correctly answer related questions. The answers in the table are produced by FinQANet(bert-base). \textbf{Gold inds} represents the tabular information about the problem filtered by the retriever, where red represents errors. \textbf{Prediction} represents the answer given by the model.}
%\resizebox{.95\columnwidth}{!}{
\begin{tabular}{l*{6}{c}}
  \toprule
  \textbf{Text Input}& \multicolumn{5}{c}{\makecell{table of contents in this form 10-k , we discuss non-gaap income before income taxes , non-gaap net income , non-gaap net \\income per diluted share , ebitda , adjusted ebitda and adjusted ebitda margin, which are non-gaap financial measures ....}}\\
  \multirow{4}*{\textbf{Table Input}} &
  -& net sales & gross profit& income from operations&...\\
  &years ended december 31 , 2017&$\$ 15191.5$&$2449.9$&$866.1$&...\\
  &years ended december 31 , 2016&...&...&...&...\\
  &...&...&...&...&...\\
  \textbf{Question}& \multicolumn{5}{c}{what was 2017 gross margin percent?}\\
  \textbf{Related Formula}& \multicolumn{5}{c}{$gross\_margins=gross\_profit/revenue$}\\
  \midrule
  \multicolumn{3}{c}{\textbf{Model Trained on FinLLMs}} & \multicolumn{3}{c}{\textbf{Model Trained on FinQA}} \\
  \textbf{Gold inds}&\multicolumn{2}{c}{\textbf{Prediction}}&\textbf{Gold inds}&\multicolumn{2}{c}{\textbf{Prediction}}\\
\makecell{net sales:$\$ 15191.5$ \\gross profit:$2449.9$}&\multicolumn{2}{c}{$divide(2449.9, 15191.5)$ \checkmark}&\makecell{\textcolor{red}{income from operations:$866.1$} \\gross profit:$2449.9$}&\multicolumn{2}{c}{$divide(\textcolor{red}{866.1}, \textcolor{red}{2449.9})$\XSolid}\\

  \bottomrule
\end{tabular}

\label{table6}
\end{table*}

\subsection{Challenges and Overlooks}
%%%%%%%%%%
Despite the success of large language model generated FinLLMs over manually labeled datasets, there exist several opening challenges and require interdisciplinary efforts and collaboration toward advancing data synthesis research in financial domain.

\textbf{Data Distribution discrepancy.} Our analysis on the FinLLMs spots a distributional discrepancy between the generated data and the original data. Despite the utilization of prompt techniques, certain issues persist. One of the primary reasons behind lies in the variance between the knowledge distribution present in the manually collected formulas and the knowledge needed for real-world scenarios. For instance, it is challenging to obtain the frequency of questions asked in real-life situations and the knowledge required to address those questions.

\textbf{Gold Statement Selection at Retriever.} Selection of statements related to the questions can also be further improved. Our current approach relies on results obtained through matching numbers found in the answers. However, this method is imprecise and can potentially affect the training effectiveness of the retriever. In our future work, we aim to address this issue by leveraging trainable models to enhance the selection process.

\textbf{Privacy Risks.} Data synthesis using large language models carries potential privacy and security risks, especially in financial domain which involves strict regulations~\cite{assefa2020generating}. This approach hinges on the extensive language capabilities of models like GPT-3.5, which, in turn, derive their knowledge from publicly available corpora used for training. Some researchers have found that machine learning models, during the training process, might memorize training data, posing a risk of privacy leakage~\cite{carlini2022privacy,carlini2021extracting,song2017machine}
The possibility of privacy breaches from AI-driven data synthesis stems from the fact that the large language model itself might pose privacy vulnerabilities. Currently, there has been a significant amount of privacy protection research in the field of data synthesis, such as data purification and differential privacy~\cite{lu2024machine}. In the meantime, current research suggests that there exists a gap between these privacy-preserving methods and their practical applicability~\cite{brown2022does}.

\textbf{Debate in AI Synthesized Data.} 
The utilization of AI-generated data as a training approach has sparked a fair share of controversy within the research community. On one hand, some researchers endorse it as the future trajectory of large model training~\cite{singh2023beyond}. These proponents argue that harnessing the generative capabilities of models like GPT-3.5 can exponentially expand training data, thereby enhancing model proficiency across various domains~\cite{he2022synthetic,tian2023learning,lu2024machine}. On the other hand, some believe that this approach carries significant risks~\cite{shumailov2023model}, as AI-generated data may introduce instability to the models themselves.  These concerns revolve around the fidelity and accuracy of the synthetic data, as well as the risk of models learning from their own biases and errors during the data generation process.

\section{Related Work}

\textbf{Questions Answering. }{ Extractive question-answering (QA) involves the task of locating answers to questions within a given text. There have been several QA datasets incorporating numerical understandings and calculations~\cite {DBLP:journals/csur/RogersGA23}. The major source is from structured tables or knowledge bases, owning the nature to succinctly organize numerical information. Popular datasets include TabFact~\cite {DBLP:conf/iclr/ChenWCZWLZW20}, DROP dataset~\cite {DBLP:conf/naacl/DuaWDSS019}, HybridQA~\cite {DBLP:conf/emnlp/ChenZCXWW20}, etc. All these existing datasets are built upon the general domain, and their numerical reasoning revolves around basic  calculations and comparisons. In contrast, our dataset focuses on the finance domain and combines both structured tables and unstructured texts, which presents a significantly more intricate landscape in terms of numerical reasoning questions. }

\textbf{Financial natural language processing (NLP). }{Financial NLP has become a major application domain attracting growing attention. Previous works in the finance domain include risk management for fraud detection~\cite {DBLP:conf/acl/HanBHDBW18}. Recently, in the context of financial reports, more complicated arithmetic expression generation has taken a vital part~\cite {MultiHiertt}. In this domain, widely recognized datasets include FinQA~\cite {chen-etal-2021-finqa} and TAT-QA~\cite {TAT-zhu-etal-2021-tat}. These datasets are unique due to their emphasis on hybrid question answering, which entails answering questions based on information presented in heterogeneous formats, including both tables and textual content. Based on FinQA, there are some works trying to extend the data set, such as MultiHiertt~\cite {MultiHiertt}. Some works try to improve the performance of the method, such as DyRRen~\cite{Li_Zhu_Liu_Ju_Qu_Cheng_2023}. There are also some studies trying to improve the TapOP (TAT-QA baseline) method to obtain better results, such as MVGE~\cite{wei2023multi} and RegHNT~\cite{lei2022answering}}.

\textbf{Financial Dataset Generation. }Financial data encompasses highly sensitive and personally identifiable information about customers. The utilization and dissemination of such data, particularly for research purposes beyond the originating organizations, are subject to stringent restrictions. One approach to address this limitation is the generation of synthetic data.
Recent advances in deep learning, such as generative adversarial networks~\cite {DBLP:journals/pvldb/ParkMGJPK18,DBLP:conf/nips/XuSCV19,DBLP:conf/acml/0001KBC21}, autoencoders~\cite {DBLP:conf/prdc/LiTH19,DBLP:conf/nips/XuSCV19,DBLP:journals/corr/abs-2105-08204}, language models~\cite {DBLP:conf/iclr/BorisovSLPK23}, and diffusion models~\cite {DBLP:conf/icml/KotelnikovBRB23} have been applied to synthetic tabular data generation. These papers demonstrate deep learning models’ capacity to produce more realistic data than traditional approaches such as Bayesian networks~\cite {DBLP:conf/nips/XuSCV19}.

Nevertheless, The existing models for generating relational datasets are primarily designed for tabular forms and are unable to generate text relevant to the tabular content, which is exceptionally crucial in financial data. The data generated by these methods perform well in terms of similarity to the original data. However, this type of data cannot be used in QA questions.

\textbf{QA Dataset Generation. }{Question Generation Systems can be categorized into two main domains: closed domain and open domain. Closed-domain question generation focuses on generating questions within a specific domain, such as medicine~\cite {wang2008automatic,shen2020generation} or educational~\cite {araki2016generating} text. In this context, the questions typically leverage domain-specific knowledge constrained by an ontology. Developing a QA system often demands a vast number of human-annotated training instances. However, obtaining such annotations from experts can be notably expensive in the financial domain. Recent studies have attempted to reduce annotation costs by generating synthetic datasets from unlabeled corpora~\cite {DBLP:conf/acl/AlbertiAPDC19,DBLP:conf/emnlp/LyuSGF0021}. Nevertheless, these efforts have been confined to  generating questions with single-span answers, leaving out a significant category of questions that necessitate answers with multiple spans~\cite{DBLP:journals/bmcbi/TsatsaronisBMPZ15}. Although questions  with multiple spans constitute a large portion of the questions asked in practice, the automatic generation of these QA datasets remains an under-explored area. FinLLMs fill a gap in the existing QA datasets, enabling a more nuanced exploration of advanced numerical reasoning scenarios. QA dataset generation frameworks typically consist of answer extraction, question generation, and model filtering, all of which hinge on human-labeled data for training. Unfortunately, this supervised approach is not applicable to list QA because existing large QA datasets contain only a few or no list-type questions \cite {chen-etal-2021-finqa}.}

\section{Conclusion}
We have presented the development of FinLLMs, a framework for financial reasoning dataset generation with large language models. Our goal is to
alleviate the data scarcity problem in this field. 
The key steps of our approach involve collecting common financial formulas along with their associated variables for graph construction. These formulas are then augmented by identifying and combining those containing the same variables as new elements. Finally, we utilize GPT-3.5 to generate financial question-answering data, comprising tables and long texts based on the formula set. Our experimental results have demonstrated
 that synthetic data generated by FinLLMs can effectively
improve the performance of several large-scale question-answering models in
the financial domain, over two benchmark financial question-answering datasets: FinQA and TAT-QA. We also analyzed the impact of each stage of data generation on the model's results. 
Our error analysis revealed certain limitations in the method, such as neglect of numerical units. As part of our future work, we will explore trainable models for filtering supporting facts, rather than relying on numerical matching alone. Additionally, we aim to expand the dataset by incorporating more advanced formulas, thereby increasing its scale and applicability.

\bibliographystyle{IEEEtran}
\bibliography{TBD}

\begin{thebibliography}{10}
\providecommand{\url}[1]{#1}
\csname url@samestyle\endcsname
\providecommand{\newblock}{\relax}
\providecommand{\bibinfo}[2]{#2}
\providecommand{\BIBentrySTDinterwordspacing}{\spaceskip=0pt\relax}
\providecommand{\BIBentryALTinterwordstretchfactor}{4}
\providecommand{\BIBentryALTinterwordspacing}{\spaceskip=\fontdimen2\font plus
\BIBentryALTinterwordstretchfactor\fontdimen3\font minus \fontdimen4\font\relax}
\providecommand{\BIBforeignlanguage}[2]{{%
\expandafter\ifx\csname l@#1\endcsname\relax
\typeout{** WARNING: IEEEtran.bst: No hyphenation pattern has been}%
\typeout{** loaded for the language `#1'. Using the pattern for}%
\typeout{** the default language instead.}%
\else
\language=\csname l@#1\endcsname
\fi
#2}}
\providecommand{\BIBdecl}{\relax}
\BIBdecl

\bibitem{wei2023multi}
Y.~Wei, F.~Lei, Y.~Zhang, J.~Zhao, and K.~Liu, ``Multi-view graph representation learning for answering hybrid numerical reasoning question,'' \emph{arXiv preprint arXiv:2305.03458}, 2023.

\bibitem{lei2022answering}
F.~Lei, S.~He, X.~Li, J.~Zhao, and K.~Liu, ``Answering numerical reasoning questions in table-text hybrid contents with graph-based encoder and tree-based decoder,'' in \emph{International Conference on Computational Linguistics}, 2022, pp. 1379--1390.

\bibitem{needles2013principles}
B.~E. Needles, M.~Powers, and S.~V. Crosson, \emph{Principles of accounting}.\hskip 1em plus 0.5em minus 0.4em\relax Cengage Learning, 2013.

\bibitem{chen-etal-2021-finqa}
Z.~Chen, W.~Chen, C.~Smiley, S.~Shah, I.~Borova, D.~Langdon, R.~Moussa, M.~Beane, T.-H. Huang, B.~Routledge, and W.~Y. Wang, ``{F}in{QA}: A dataset of numerical reasoning over financial data,'' in \emph{Conference on Empirical Methods in Natural Language Processing}, 2021, pp. 3697--3711.

\bibitem{TAT-zhu-etal-2021-tat}
F.~Zhu, W.~Lei, Y.~Huang, C.~Wang, S.~Zhang, J.~Lv, F.~Feng, and T.-S. Chua, ``{TAT}-{QA}: A question answering benchmark on a hybrid of tabular and textual content in finance,'' in \emph{Annual Meeting of the Association for Computational Linguistics}, 2021, pp. 3277--3287.

\bibitem{Li_Zhu_Liu_Ju_Qu_Cheng_2023}
X.~Li, Y.~Zhu, S.~Liu, J.~Ju, Y.~Qu, and G.~Cheng, ``Dyrren: A dynamic retriever-reranker-generator model for numerical reasoning over tabular and textual data,'' \emph{AAAI Conference on Artificial Intelligence}, vol.~37, no.~11, pp. 13\,139--13\,147, Jun. 2023.

\bibitem{DBLP:conf/naacl/DevlinCLT19}
J.~Devlin, M.~Chang, K.~Lee, and K.~Toutanova, ``{BERT:} pre-training of deep bidirectional transformers for language understanding,'' in \emph{Conference of the North American Chapter of the Association for Computational Linguistics: Human Language Technologies}, 2019, pp. 4171--4186.

\bibitem{DBLP:journals/corr/abs-1907-11692}
Y.~Liu, M.~Ott, N.~Goyal, J.~Du, M.~Joshi, D.~Chen, O.~Levy, M.~Lewis, L.~Zettlemoyer, and V.~Stoyanov, ``Roberta: {A} robustly optimized {BERT} pretraining approach,'' \emph{CoRR}, vol. abs/1907.11692, 2019.

\bibitem{MultiHiertt}
Y.~Zhao, Y.~Li, C.~Li, and R.~Zhang, ``Multihiertt: Numerical reasoning over multi hierarchical tabular and textual data,'' in \emph{Annual Meeting of the Association for Computational Linguistics}, 2022, pp. 6588--6600.

\bibitem{DBLP:conf/aaai/LeeKK23}
S.~Lee, H.~Kim, and J.~Kang, ``{LIQUID:} {A} framework for list question answering dataset generation,'' in \emph{{AAAI} Conference on Artificial Intelligence}, 2023, pp. 13\,014--13\,024.

\bibitem{DBLP:conf/naacl/DuaWDSS019}
D.~Dua, Y.~Wang, P.~Dasigi, G.~Stanovsky, S.~Singh, and M.~Gardner, ``{DROP:} {A} reading comprehension benchmark requiring discrete reasoning over paragraphs,'' in \emph{Conference of the North American Chapter of the Association for Computational Linguistics: Human Language Technologies}.\hskip 1em plus 0.5em minus 0.4em\relax Association for Computational Linguistics, 2019, pp. 2368--2378.

\bibitem{DBLP:conf/acl/HanBHDBW18}
J.~Han, U.~Barman, J.~Hayes, J.~Du, E.~Burgin, and D.~Wan, ``Nextgen {AML:} distributed deep learning based language technologies to augment anti money laundering investigation,'' in \emph{Conference of the Association for Computational Linguistics}.\hskip 1em plus 0.5em minus 0.4em\relax Association for Computational Linguistics, 2018, pp. 37--42.

\bibitem{DBLP:conf/acl/AlbertiAPDC19}
C.~Alberti, D.~Andor, E.~Pitler, J.~Devlin, and M.~Collins, ``Synthetic {QA} corpora generation with roundtrip consistency,'' in \emph{Conference of the Association for Computational Linguistics}.\hskip 1em plus 0.5em minus 0.4em\relax Association for Computational Linguistics, 2019, pp. 6168--6173.

\bibitem{DBLP:conf/emnlp/LyuSGF0021}
C.~Lyu, L.~Shang, Y.~Graham, J.~Foster, X.~Jiang, and Q.~Liu, ``Improving unsupervised question answering via summarization-informed question generation,'' in \emph{Conference on Empirical Methods in Natural Language Processing}.\hskip 1em plus 0.5em minus 0.4em\relax Association for Computational Linguistics, 2021, pp. 4134--4148.

\bibitem{DBLP:journals/bmcbi/TsatsaronisBMPZ15}
G.~Tsatsaronis, G.~Balikas, P.~Malakasiotis, I.~Partalas, M.~Zschunke, M.~R. Alvers, D.~Weissenborn, A.~Krithara, S.~Petridis, D.~Polychronopoulos, Y.~Almirantis, J.~Pavlopoulos, N.~Baskiotis, P.~Gallinari, T.~Arti{\`{e}}res, A.~N. Ngomo, N.~Heino, {\'{E}}.~Gaussier, L.~Barrio{-}Alvers, M.~Schroeder, I.~Androutsopoulos, and G.~Paliouras, ``An overview of the {BIOASQ} large-scale biomedical semantic indexing and question answering competition,'' \emph{{BMC} Bioinform.}, vol.~16, pp. 138:1--138:28, 2015.

\bibitem{DBLP:conf/iclr/ChenWCZWLZW20}
W.~Chen, H.~Wang, J.~Chen, Y.~Zhang, H.~Wang, S.~Li, X.~Zhou, and W.~Y. Wang, ``Tabfact: {A} large-scale dataset for table-based fact verification,'' in \emph{International Conference on Learning Representations}, 2020.

\bibitem{DBLP:conf/emnlp/ChenZCXWW20}
W.~Chen, H.~Zha, Z.~Chen, W.~Xiong, H.~Wang, and W.~Y. Wang, ``Hybridqa: {A} dataset of multi-hop question answering over tabular and textual data,'' in \emph{Findings of the Association for Computational Linguistics: {EMNLP}}.\hskip 1em plus 0.5em minus 0.4em\relax Association for Computational Linguistics, 2020, pp. 1026--1036.

\bibitem{DBLP:journals/csur/RogersGA23}
A.~Rogers, M.~Gardner, and I.~Augenstein, ``{QA} dataset explosion: {A} taxonomy of {NLP} resources for question answering and reading comprehension,'' \emph{{ACM} Computing Surveys}, vol.~55, no.~10, pp. 197:1--197:45, 2023.

\bibitem{DBLP:conf/acl/ZhaoLLZ22}
Y.~Zhao, Y.~Li, C.~Li, and R.~Zhang, ``Multihiertt: Numerical reasoning over multi hierarchical tabular and textual data,'' in \emph{Annual Meeting of the Association for Computational Linguistics}, 2022, pp. 6588--6600.

\bibitem{DBLP:journals/pvldb/ParkMGJPK18}
N.~Park, M.~Mohammadi, K.~Gorde, S.~Jajodia, H.~Park, and Y.~Kim, ``Data synthesis based on generative adversarial networks,'' \emph{{VLDB} Endowment}, vol.~11, no.~10, pp. 1071--1083, 2018.

\bibitem{DBLP:conf/nips/XuSCV19}
L.~Xu, M.~Skoularidou, A.~Cuesta{-}Infante, and K.~Veeramachaneni, ``Modeling tabular data using conditional {GAN},'' in \emph{Conference on Neural Information Processing Systems}, 2019, pp. 7333--7343.

\bibitem{DBLP:conf/acml/0001KBC21}
Z.~Zhao, A.~Kunar, R.~Birke, and L.~Y. Chen, ``{CTAB-GAN:} effective table data synthesizing,'' in \emph{Asian Conference on Machine Learning}.\hskip 1em plus 0.5em minus 0.4em\relax {PMLR}, 2021, pp. 97--112.

\bibitem{DBLP:conf/prdc/LiTH19}
S.~Li, B.~Tai, and Y.~Huang, ``Evaluating variational autoencoder as a private data release mechanism for tabular data,'' in \emph{{IEEE} Pacific Rim International Symposium on Dependable Computing}, 2019, pp. 198--206.

\bibitem{DBLP:journals/corr/abs-2105-08204}
S.~Darabi and Y.~Elor, ``Synthesising multi-modal minority samples for tabular data,'' \emph{CoRR}, vol. abs/2105.08204, 2021.

\bibitem{DBLP:conf/iclr/BorisovSLPK23}
V.~Borisov, K.~Se{\ss}ler, T.~Leemann, M.~Pawelczyk, and G.~Kasneci, ``Language models are realistic tabular data generators,'' in \emph{International Conference on Learning Representations}, 2023.

\bibitem{tian2023learning}
Y.~Tian, L.~Fan, K.~Chen, D.~Katabi, D.~Krishnan, and P.~Isola, ``Learning vision from models rivals learning vision from data,'' \emph{arXiv preprint arXiv:2312.17742}, 2023.

\bibitem{lu2024machine}
Y.~Lu, M.~Shen, H.~Wang, X.~Wang, C.~van Rechem, and W.~Wei, ``Machine learning for synthetic data generation: A review,'' \emph{arXiv preprint arXiv:2302.04062}, 2023.

\bibitem{he2022synthetic}
R.~He, S.~Sun, X.~Yu, C.~Xue, W.~Zhang, P.~Torr, S.~Bai, and X.~QI, ``Is synthetic data from generative models ready for image recognition?'' in \emph{International Conference on Learning Representations}, 2022.

\bibitem{DBLP:conf/icml/KotelnikovBRB23}
A.~Kotelnikov, D.~Baranchuk, I.~Rubachev, and A.~Babenko, ``Tabddpm: Modelling tabular data with diffusion models,'' in \emph{International Conference on Machine Learning}, vol. 202.\hskip 1em plus 0.5em minus 0.4em\relax {PMLR}, 2023, pp. 17\,564--17\,579.

\bibitem{wang2008automatic}
W.~Wang, T.~Hao, and W.~Liu, ``Automatic question generation for learning evaluation in medicine,'' in \emph{Advances in Web Based Learning-ICWL International Conference}, vol. 4823.\hskip 1em plus 0.5em minus 0.4em\relax Springer Science \& Business Media, 2008, p. 242.

\bibitem{shen2020generation}
S.~Shen, Y.~Li, N.~Du, X.~Wu, Y.~Xie, S.~Ge, T.~Yang, K.~Wang, X.~Liang, and W.~Fan, ``On the generation of medical question-answer pairs,'' in \emph{AAAI Conference on Artificial Intelligence}, vol.~34, no.~05, 2020, pp. 8822--8829.

\bibitem{carlini2022privacy}
N.~Carlini, M.~Jagielski, C.~Zhang, N.~Papernot, A.~Terzis, and F.~Tramer, ``The privacy onion effect: Memorization is relative,'' \emph{Conference on Neural Information Processing Systems}, vol.~35, pp. 13\,263--13\,276, 2022.

\bibitem{chen-etal-2022-convfinqa}
Z.~Chen, S.~Li, C.~Smiley, Z.~Ma, S.~Shah, and W.~Y. Wang, ``{C}onv{F}in{QA}: Exploring the chain of numerical reasoning in conversational finance question answering,'' in \emph{Conference on Empirical Methods in Natural Language Processing}, 2022, pp. 6279--6292.

\bibitem{araki2016generating}
J.~Araki, D.~Rajagopal, S.~Sankaranarayanan, S.~Holm, Y.~Yamakawa, and T.~Mitamura, ``Generating questions and multiple-choice answers using semantic analysis of texts,'' in \emph{International Conference on Computational Linguistics}, 2016, pp. 1125--1136.

\bibitem{zhu2023soargraph}
F.~Zhu, M.~Li, J.~Xiao, F.~Feng, C.~Wang, and T.~S. Chua, ``Soargraph: Numerical reasoning over financial table-text data via semantic-oriented hierarchical graphs,'' in \emph{Companion Proceedings of the ACM Web Conference 2023}, 2023, pp. 1236--1244.

\bibitem{singh2023beyond}
A.~Singh, J.~D. Co-Reyes, R.~Agarwal, A.~Anand, P.~Patil, P.~J. Liu, J.~Harrison, J.~Lee, K.~Xu, A.~Parisi \emph{et~al.}, ``Beyond human data: Scaling self-training for problem-solving with language models,'' \emph{arXiv preprint arXiv:2312.06585}, 2023.

\bibitem{brown2022does}
H.~Brown, K.~Lee, F.~Mireshghallah, R.~Shokri, and F.~Tram{\`e}r, ``What does it mean for a language model to preserve privacy?'' in \emph{ACM Conference on Fairness, Accountability, and Transparency}, 2022, pp. 2280--2292.

\bibitem{carlini2021extracting}
N.~Carlini, F.~Tramer, E.~Wallace, M.~Jagielski, A.~Herbert-Voss, K.~Lee, A.~Roberts, T.~Brown, D.~Song, U.~Erlingsson \emph{et~al.}, ``Extracting training data from large language models,'' in \emph{USENIX Security Symposium}, 2021, pp. 2633--2650.

\bibitem{song2017machine}
C.~Song, T.~Ristenpart, and V.~Shmatikov, ``Machine learning models that remember too much,'' in \emph{ACM SIGSAC Conference on computer and communications security}, 2017, pp. 587--601.

\bibitem{assefa2020generating}
S.~A. Assefa, D.~Dervovic, M.~Mahfouz, R.~E. Tillman, P.~Reddy, and M.~Veloso, ``Generating synthetic data in finance: opportunities, challenges and pitfalls,'' in \emph{ACM International Conference on AI in Finance}, 2020, pp. 1--8.

\bibitem{shumailov2023model}
I.~Shumailov, Z.~Shumaylov, Y.~Zhao, Y.~Gal, N.~Papernot, and R.~Anderson, ``Model dementia: Generated data makes models forget,'' \emph{arXiv e-prints}, pp. arXiv--2305, 2023.

\end{thebibliography}

\vspace{-15pt} 

\begin{IEEEbiography}
[{\includegraphics[width=1in,height=1.25in,clip,keepaspectratio]{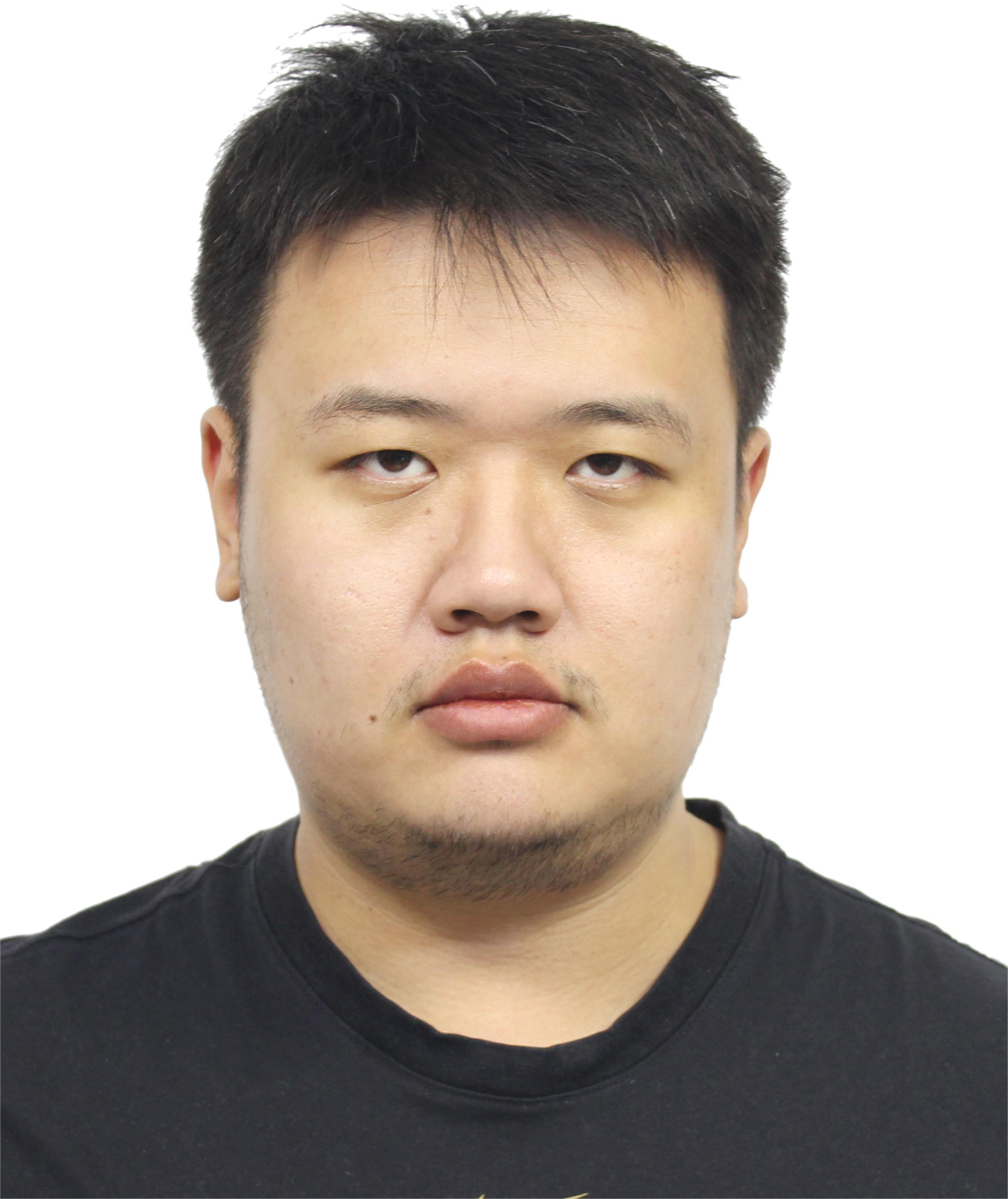}}]
{Ziqiang Yuan}  received the BE and ME degrees in Computer Science from Beijing Institute of Technology. He is currently working toward the PhD degree with the Beijing Institute of Technology. His primary research interests include Natural Language Processing, QA Systems, and prompt Learning.
\end{IEEEbiography} 

\vspace{-15pt} 

\begin{IEEEbiography}
[{\includegraphics[width=1in,height=1.25in,clip,keepaspectratio]{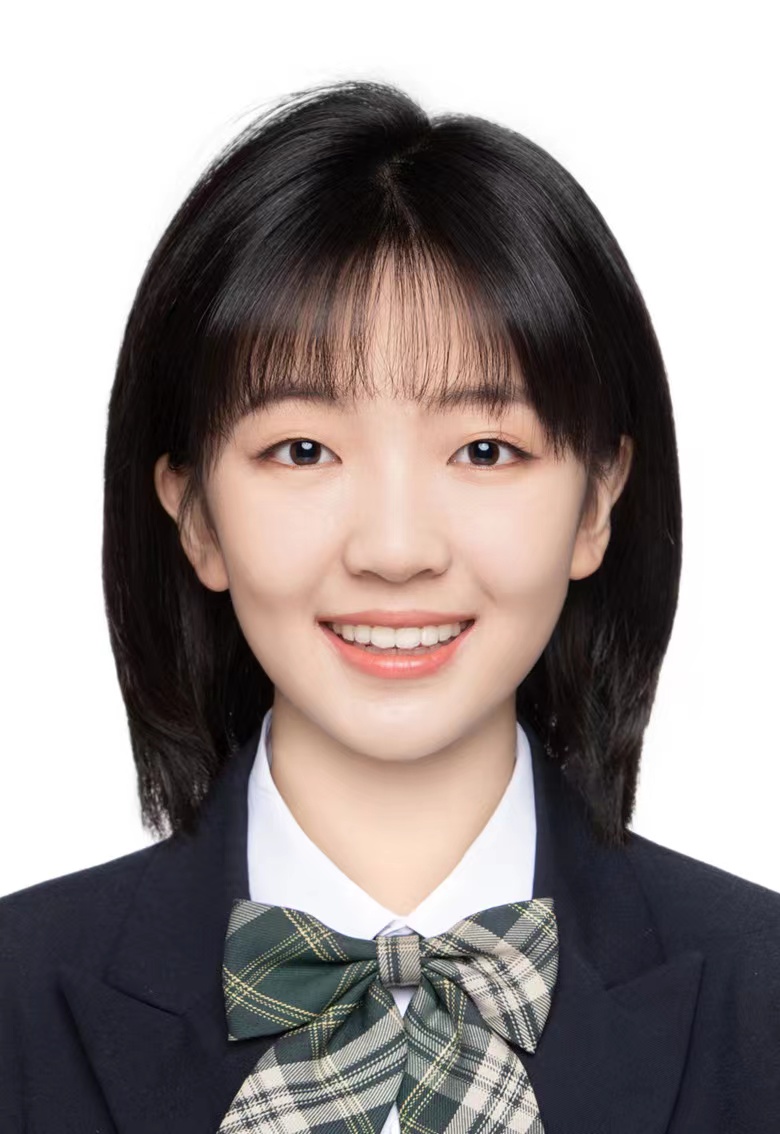}}]
{Kaiyuan Wang} Kaiyuan Wang is currently a sophomore at Central University of Finance and Economics (CUFE).   She became interested in intelligent financial systems during her studies. Her research interests include research on Automatic Generation of Question and Answer Pairs Based on Tensor Graph of Accounting Knowledge. 
\end{IEEEbiography} 

\vspace{-15pt} 

\begin{IEEEbiography}
[{\includegraphics[width=1in,height=1.25in,clip,keepaspectratio]{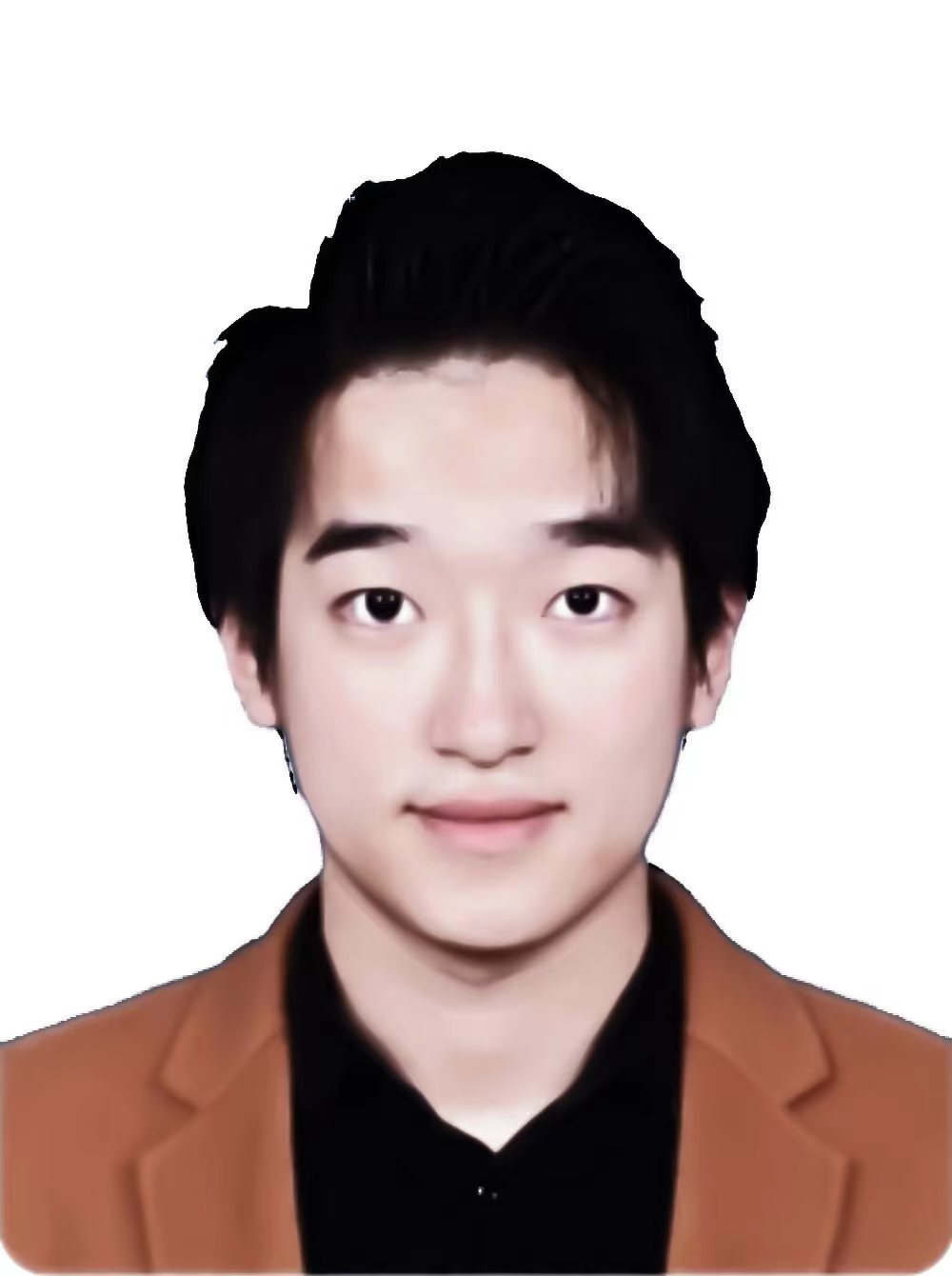}}]
{Shoutai Zhu} is currently a master's degree student at the School of Computer Science at Beijing Institute of Technology, he received a bachelor's degree in computer science and technology from Beijing Institute of Technology in 2022. His main research interests include Natural Language Processing, Reinforcement Learning, and Prompt Learning.
\end{IEEEbiography} 

\vspace{-15pt} 

\begin{IEEEbiography}
[{\includegraphics[width=1in,height=1.25in,clip,keepaspectratio]{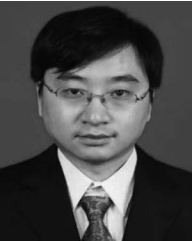}}]
{Ye Yuan} 
received the BE, ME, and PhD degrees in computer science from Northeastern University, in
2004, 2007, and 2011, respectively. Currently, he is a professor with the Department of Computer Science, Beijing Institute of Technology, China. His research interests include graph databases, probabilistic databases, and social network analysis.
\end{IEEEbiography}

\vspace{-15pt} 
\begin{IEEEbiography}
[{\includegraphics[width=1in,height=1.25in,clip,keepaspectratio]{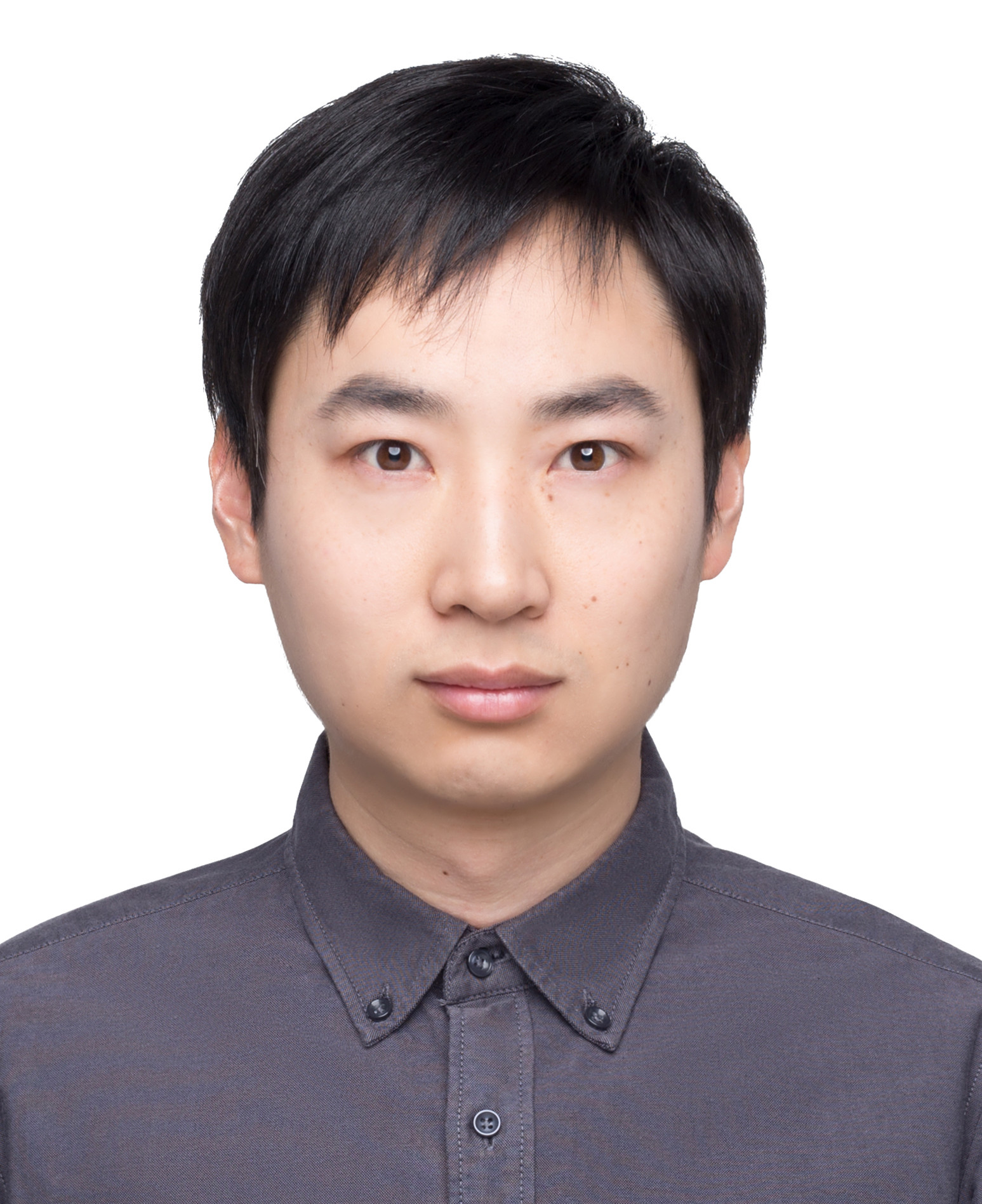}}]
{Jingya Zhou} received the BE degree in computer science from Anhui Normal University, Wuhu, in 2005, and his PhD degree in computer science from Southeast University, Nanjing, in 2013. He is currently a professor with the School of Computer Science and Technology, Soochow University, Suzhou, China. His research interests include cloud and edge computing, network embedding, data mining, online social networks and recommender systems.
\end{IEEEbiography}

\vspace{-15pt} 

\begin{IEEEbiography}
[{\includegraphics[width=1in,height=1.25in,clip,keepaspectratio]{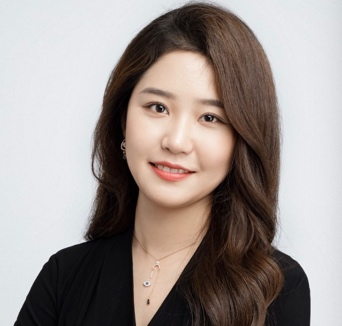}}]
{Yanlin Zhu} received her Master of Arts in Mathematics of Finance from Columbia University in 2021. She is currently a Financial Engineer at Moody’s Analytics in New York City.
\end{IEEEbiography}

\vspace{-15pt} 

\begin{IEEEbiography}
[{\includegraphics[width=1in,height=1.25in,clip,keepaspectratio]{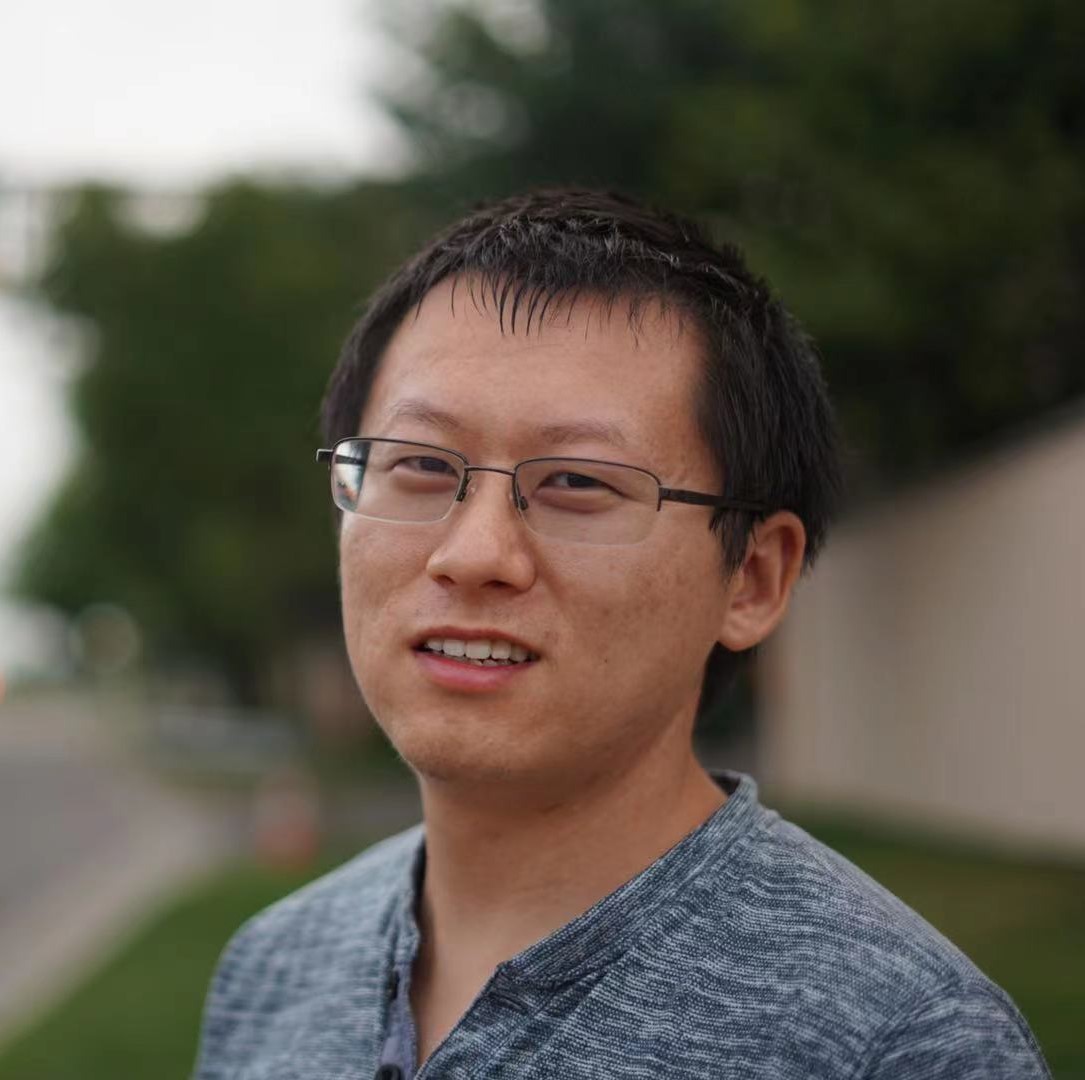}}]
{Wenqi Wei} 
is currently a tenure-track assistant professor in the Computer and Information Sciences Department, Fordham University. He 
obtained his PhD in the School of Computer Science, Georgia Institute of Technology. %where he is advised by Prof.~Ling Liu. 
He received his B.E. degree from the School of Electronic Information
and Communications, Huazhong University of Science
and Technology. His research interests include data privacy, machine learning security, and big data analytics. 
\end{IEEEbiography}

\vfill

\end{document}